\documentclass[preprint,10pt]{elsarticle} 
\usepackage{float}
\usepackage{amsmath}
\usepackage{graphicx}
\usepackage[hidelinks]{hyperref}
\usepackage[utf8]{inputenc}
\usepackage{times} 
\usepackage[verbose,tmargin=1in,bmargin=1in,lmargin=1in,rmargin=1in]{geometry}
\usepackage{booktabs} 
\usepackage{setspace} 
\usepackage{lipsum} 
\usepackage{graphicx} 
\usepackage{url} 
\usepackage{siunitx} 
\usepackage{subfiles} 
\usepackage{float}  
\usepackage{stfloats} 
\usepackage{threeparttable} 
\usepackage{tablefootnote} 
\usepackage{nameref} 
\usepackage{amsmath} 
\usepackage{multirow} 
\usepackage{multicol}
\usepackage[table]{xcolor}
\usepackage{colortbl}
\usepackage{pdfpages} 
\usepackage{comment}
\usepackage{pifont} 
\usepackage{tcolorbox}
\usepackage{subcaption}
\usepackage{array}
\usepackage[acronym]{glossaries}
\glsdisablehyper 

\newcommand{\cmark}{\ding{51}}

\newcommand{\figsup}[2]{\label{#1}}
\newcommand{\tabsup}[2]{\label{#1}}
\newcommand{\secsup}[2]{\label{#1}}

\usepackage{xcolor}
\usepackage{titlesec}
\titleformat*{\section}{\large\bfseries\sffamily}
\titleformat*{\subsection}{\normalsize\bfseries\sffamily}
\titleformat*{\subsubsection}{\small\bfseries\sffamily}
\newcommand{\absdiv}[1]{%
  \par\addvspace{.5\baselineskip}
  \noindent\textbf{#1}\quad\ignorespaces}
\renewenvironment{abstract}{\global\setbox\absbox=\vbox\bgroup
  \hsize=\textwidth%
  \noindent\unskip\textbf{\large Abstract}
  \par\medskip\noindent\unskip\ignorespaces}{\egroup}
\makeatletter
\renewcommand\@biblabel[1]{#1} 
\makeatother

\biboptions{sort&compress}
\biboptions{super}

\journal{Journal}

\usepackage{titlesec}
\titleformat{\section}[block]{\normalfont\Large\bfseries}{}{0em}{}
\titleformat{\subsection}[block]{\normalfont\large\bfseries}{}{0em}{}
\titleformat{\subsubsection}[block]{\normalfont\normalsize\bfseries}{}{0em}{}

\makeatletter
\let\@afterindenttrue\@afterindentfalse
\def\ps@pprintTitle{%
  \let\@oddhead\@empty
  \let\@evenhead\@empty
  \def\@oddfoot{\footnotesize Under review.\hfill}%
  \let\@evenfoot\@oddfoot
}
\makeatother
\newacronym{ai}{AI}{\emph{artificial intelligence}}
\newacronym{auprc}{AUPRC}{\emph{ area under the precision-recall curve}}
\newacronym{auroc}{AUROC}{\emph{area under the receiver operating characteristic curve}}
\newacronym{bs}{BS}{Brier score}
\newacronym{cap}{CAP}{\emph{cost-aware prediction}}
\newacronym{cdss}{CDSS}{\emph{clinical decision support system}}
\newacronym{ci}{CI}{confidence interval}
\newacronym{cip}{CIP}{\emph{clinical impact projection}}
\newacronym{dca}{DCA}{\emph{decision curve analysis}}
\newacronym{lgb}{LGB}{\emph{light gradient boosting machine}}
\newacronym{llm}{LLM}{\emph{large language model}}
\newacronym{lr}{LR}{\emph{logistic regression}}
\newacronym{mct}{MCT}{\emph{misclassification cost term}}
\newacronym{ml}{ML}{\emph{machine learning}}
\newacronym{qol}{QoL}{\emph{quality of life}}
\newacronym{rf}{RF}{\emph{random forest}}
\newacronym{xgb}{XGB}{\emph{eXtreme gradient boosting}}
\newacronym{fn}{FN}{\emph{false negative}}
\newacronym{fp}{FP}{\emph{false positive}}
\newacronym{tn}{TN}{\emph{true negative}}
\newacronym{tp}{TP}{\emph{true positive}}

\begin{document}

\begin{frontmatter}
\title{Cost-Aware Prediction (CAP): An LLM-Enhanced Machine Learning Pipeline and Decision Support System for Heart Failure Mortality Prediction}

\author{
    Yinan Yu\corref{corrauth}\textsuperscript{1}
}
\author{
    Falk Dippel\textsuperscript{2}
}
\author{
    Christina E. Lundberg\textsuperscript{3}\textsuperscript{,}\textsuperscript{5}
}
\author{
    Martin Lindgren\textsuperscript{3}\textsuperscript{,}\textsuperscript{4}
}
\author{
    \\Annika Rosengren\textsuperscript{3}\textsuperscript{,}\textsuperscript{4}
}
\author{
    Martin Adiels\textsuperscript{6}
}
\author{
    Helen Sjöland\textsuperscript{3}\textsuperscript{,}\textsuperscript{4}
}

\address{\textsuperscript{1}Department of Computer Science and Engineering, Chalmers University of Technology and University of Gothenburg, Gothenburg, Sweden}
\address{\textsuperscript{2}Sahlgrenska University Hospital, Gothenburg, Sweden}
\address{\textsuperscript{3}Department of Molecular and Clinical Medicine, Sahlgrenska Academy, University of Gothenburg, Gothenburg, Sweden}
\address{\textsuperscript{4}Department of Medicine, Geriatrics and Emergency Medicine, Sahlgrenska University Hospital, Gothenburg, Sweden}
\address{\textsuperscript{5}Department of Food and Nutrition, and Sport Science, Faculty of Education, University of Gothenburg, Gothenburg, Sweden}
\address{\textsuperscript{6}School of Public Health and Community Medicine, Institute of Medicine, University of Gothenburg, Gothenburg, Sweden}

\cortext[corrauth]{Corresponding author: Yinan Yu, \href{mailto:yinan@chalmers.se}{yinan@chalmers.se}}

\begin{abstract}
  \absdiv{{Objective}}
    \Gls{ml} predictive models are often developed without considering downstream value trade-offs and clinical interpretability. This paper introduces a \gls{cap} framework that combines cost–benefit analysis assisted by \gls{llm} agents to communicate the trade-offs involved in applying \gls{ml} predictions. 

  \absdiv{{Materials and Methods}}
    We developed an \gls{ml} model predicting 1-year mortality in patients with heart failure ($N=30,021$, 22$\%$ mortality) to identify those eligible for home care. We then introduced \gls{cip} curves to visualize important cost dimensions – quality of life and healthcare provider expenses, further divided into treatment and error costs, to assess the clinical consequences of predictions. Finally, we used four \gls{llm} agents to generate patient-specific descriptions. The system was evaluated by clinicians for its decision support value. 

  \absdiv{{Results}}
    The \gls{xgb} model achieved the best performance, with an \gls{auroc} of \mbox{0.804 ($95\%$\;\text{\gls{ci}}\;$0.792\text{--}0.816$)}, \gls{auprc} of \mbox{0.529 ($95\%\;\text{\gls{ci}}\;0.502\text{--}0.558$)} and a Brier score \mbox{of 0.135 ($95\%\;\text{\gls{ci}}\;0.130\text{--}0.140$)}.  

  \absdiv{{Discussion}}
    The \gls{cip} cost curves provided a \textit{population-level} overview of cost composition across decision thresholds, whereas \gls{llm}-generated cost-benefit analysis at \textit{individual patient-levels}. The system was well received according to the evaluation by clinicians. However, feedback emphasizes the need to strengthen the technical accuracy for speculative tasks. 

  \absdiv{{Conclusion}}
    \gls{cap} utilizes \gls{llm} agents to integrate \gls{ml} classifier outcomes and cost-benefit analysis for more transparent and interpretable decision support. 

\end{abstract}

\begin{keyword}
Cost-Benefit Analysis; Clinical Decision Support Systems; Large Language Models; Heart Failure; Predictive Learning Model
\end{keyword}

\end{frontmatter}

\glsresetall

\section{Introduction}

The use of clinical prediction models based on \gls{ai} holds high promise in medicine, but transition to clinical practice remains challenging. Clinical decision-making seeks to balance available patient information to optimise favourable outcomes and minimise harm. Likewise, \glspl{cdss} typically rest on multi-objective optimisations as clinical information is computationally combined and translated to finding a local minimum of risk based on the variables included. 

Implementation of \gls{cdss} ultimately depends on satisfying regulatory requirements and earning clinicians´ trust by presenting interpretable \gls{ai} outputs, communicating the relation between competing objectives, for example, minimising cost while maximising performance and safety \cite{eu-2024/1689}. 

One method for achieving understandable \gls{ai} is to present the predicted outcome to the clinician using a decision curve to illustrate benefit relative to risk \cite{vickers_decision_2006}. A decision curve will describe the theoretical risk prediction for an undesirable outcome in an individual patient in the presence or absence of an intervention recommended by the \gls{cdss} and at various levels of probability. 

\textit{Heart failure} is a prevalent condition among older individuals \cite{wideqvist_ten_2022}, that drives substantial healthcare costs through recurrent hospitalisations, despite the fact that the condition can often successfully be treated at home \cite{desai_rehospitalization_2012, le_hospital_2022, geng_preferences_2024}. Here, we report the design of a \gls{ml} prediction model for mortality in patients with severe heart failure, aimed at supporting care-level recommendations -- either customised home care as an intervention or hospital readmission as standard care -- during periods of worsening health. Beyond predicting 1-year all-cause mortality, we proceed to address downstream consequences such as patient's \gls{qol} and healthcare resource use through the development of an \emph{interpretable} and \emph{cost-aware} predictive framework, called \gls{cap}. A semi-quantitative measure is employed to estimate patient-centred costs (e.g. patient outcomes and \gls{qol}) and costs for the healthcare provider (e.g. resource allocation and monetary expenses) as examples. The \gls{cap} framework illustrates how outcomes can be optimised under varying conditions by balancing these cost dimensions, while delivering high-accuracy predictions and interpretable decision curves. Finally, we attempt to explain the underlying calculations by user-targeted natural language explanations to facilitate interpretation and to support informed clinical decisions. 

\section{Methods} \label{sec:methods}

We introduce a decision-support framework, \glsentrylong{cap}~(\gls{cap}) that integrates three components into an interpretable system to support informed decision-making in clinical practice. 

\subsection{Dataset and prediction outcome}

\begin{figure}[!hbt]
    \centering
    \includegraphics[width=0.5\linewidth]{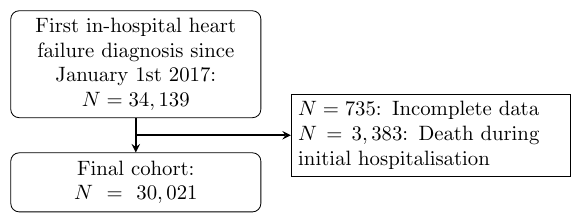}
    \caption{\textbf{Cohort flow chart indicating selection process}}
    \label{fig:flow_chart}
\end{figure}

This study includes patients $(N=34,139)$, 18 years and older, with a first in-hospital heart failure diagnosis between January 1, 2017 and December 31, 2023. Diagnoses were registered by codes I110, I130, I132, any I42 or any I50 in any position, according to the International Classification of Diseases (10th revision) (ICD-10). Patient data was collected from a comprehensive database of \emph{electronic health records} (EHRs) covering all hospital-admissions at emergency care providers in the \emph{Region Västra Götaland} (VGR) of Sweden between January 1st 2014 and December 31th 2023 to allow for 3 years of prior history. Sweden has universal healthcare providing low-cost hospital care to all residents. After exclusion of patients with missing information $(N=735)$, and patients who died during the initial hospitalisation $(N=3,383)$, the final cohort consisted of $N=30,021$ patients (Fig.~\ref{fig:flow_chart}). 

Clinical variables from patient data included: age, sex, body mass index, length of stay, comorbidities along with previous surgeries or interventions within the last 3 years, prescribed drugs within the last year prior to hospitalisation, vital signs and laboratory tests during hospitalisation, and physical mobility extracted from health records within the last 30 days before diagnostic date. For multiple measurements of the same vital sign or laboratory test, the average value was used for the respective hospitalisation. In total, 75 clinical variables were extracted from EHRs with variable-specific information, such as ICD-10 codes used for comorbidities  and Anatomical Therapeutic Chemical (ATC) codes for medications, provided in Supplementary Table~\hyperref[stab:data_collection]{S1}. 

The prediction outcome of the \gls{ml} model was all-cause mortality within 1 year after discharge after receiving a first heart failure diagnosis during hospitalisation. 

\subsubsection[\gls{ml} model]{Component 1: Supervised \gls{ml} model} 
The first component is a supervised \gls{ml} predictive model that produces a risk score between 0 and 1 for every patient. This score represents the likelihood that an event will occur. To turn this score into a yes-or-no decision, a threshold must be chosen (for example,  a decision may be made that patients with a score above 0.7 will be classified as high risk). 

\subsubsection{Component 2: Visualisation of clinical impact factors}
\label{sec:cip}
In \gls{ml}, a classification result can be \gls{tp} (a patient correctly predicted to have a high probability of mortality within one year, and who in fact does die within that time), a \gls{fp} (a patient predicted to have a high probability of mortality within one year, but who actually survives beyond that time), a \gls{tn} (a patient correctly predicted to have a low probability of mortality, and who indeed survives beyond the year), or a \gls{fn} (a patient predicted to have a low probability of mortality, but who dies within the year). 
It is commonly assumed that improved predictive accuracy automatically translates into greater clinical benefits, often without explicitly accounting for the associated clinical impact. 
This assumption implicitly categorises \glspl{fp} and \glspl{fn} as universally ``bad'', while treating \glspl{tp} and \glspl{tn} as inherently ``good''. 
This view overlooks that in clinical practice, risks can arise even from accurate predictions. 
Such risks may be influenced by clinical actions triggered by the model's outputs. Thus, while predictive accuracy remains critical, understanding the downstream impact of \gls{ml}-based predictions is essential for improving interpretability. 

To address this, we introduce the \gls{cip} curve, where the term \emph{cost} represents the clinical and organisational impact of decisions made based on model predictions. Our design of \gls{cip} considers two clinically relevant dimensions: 
\begin{itemize}
\item {\gls{qol} cost:} typically captures the patient's lived experience, including aspects relevant to their well-being during care. Lower values indicate better \gls{qol}, such as when a patient receives care that is appropriate and minimally intrusive in a familiar environment. A negative \gls{qol} cost indicates increased \gls{qol} compared to the baseline. 

\item{Healthcare system cost:} Reflects the impact on healthcare resources, including hospital bed occupancy, staff workload, and financial expenditures. Lower values indicate more efficient resource utilisation and reduced strain on healthcare systems. 
\end{itemize}

Furthermore, for each dimension, \gls{cip} considers two types of costs: clinical treatment cost and classification error cost, reflecting the clinical consequences and classification impact, respectively. More detailed definition and calculation of \gls{cip} can be found in Supplementary Section~\hyperref[ssec:cip_implementation]{CIP implementation details}.
Note that users of \gls{cip} can define additional cost dimensions and types specific for their context, and that our application represents an example of prioritised costs.

\subsubsection[Component 3: Four \gls{llm} agents]{Component 3: Cost-benefit analysis using large language model}
\label{sec:component3}
Clinicians often find \gls{ml} model outputs difficult to interpret, particularly when clinical trade-offs must be considered \cite{sanchez2022machine, mlodzinski2023assessing, hou2025explainable}. Language models can complement \gls{ml} decision support by explaining uncertainty and contextualising risks based on patient data and clinical priorities. 
To address this, the final component of the framework integrates four \gls{llm} agents to support interpretation of model outputs and associated cost–benefit trade-offs. Each agent addresses a key clinical question. The agents operate in a prompt-only mode, prioritising simplicity and traceability over integration with external tools.

\noindent\textbf{Agent I. How certain is the risk prediction?} Summarises prediction reliability for an individual patient, based on their risk score, decision threshold, and local model performance. 

\noindent\textbf{Agent II. How do I interpret the \gls{cip} cost curves?} Explains key contributors to patient-level cost, using treatment and error costs across \gls{qol} and healthcare system dimensions. 

\noindent\textbf{Agent III. How can prediction uncertainty be reduced?} Suggests actions (e.g. additional tests, record review) that may improve prediction certainty. 

\noindent\textbf{Agent IV. How can future risks be mitigated?} Provides forward-looking guidance based on predicted outcomes and potential care pathways. 

To evaluate the system, we conducted two complementary expert reviews. One clinician, involved in methods development, performed a structured assessment of the outputs for each agent (\emph{clinical development review}), focusing on accuracy and relevance. In addition, two external clinicians (experienced practising specialists in emergency medicine and cardiology, respectively) conducted a \emph{clinical user review}, evaluating interpretability, reliability, and usability of the agents’ outputs through structured questions and open-ended feedback (Supplementary Section~\hyperref[ssec:agent_evaluation]{LLM-agent evaluation details} and Supplementary Table~\hyperref[stab:cap_questionnaire]{S6}-\hyperref[stab:cap_category_feedback]{S8}). 

\subsection[]{Implementation of the \gls{cap} framework}
\label{sec:pipeline}
Here, we will illustrate how to implement this pipeline applying our use case of eligibility for home care programs through 1-year mortality prediction.   
\subsubsection[\gls{cap} Step 1]{Step 1: \gls{ml} classification model development} \label{sec:step1}

The first step is to develop and select the best performing model based on the \gls{ml} metric; area under the precision-recall curve and decision threshold based on the best F1 score (Supplementary Section~\hyperref[ssec:ml_metrics]{Machine learning model metrics} and \hyperref[ssec:ml_implementation]{Machine learning implementation details}). 

\subsubsection[\gls{cap} Step 2]{Step 2: Population-level clinical impact visualisation} \label{sec:step2}
Before generating the visualisation (i.e., the \gls{cip} cost curves), clinicians must first define the cost structure. 
Costs need to be specified separately for each type, clinical dimension, and prediction outcome scenario (\gls{tp}, \gls{fp}, \gls{tn}, and \gls{fn}). 
In our case study, the costs (ranging from $-$1 to 1), were defined based on the following assumptions, which represent a {\emph{clinical example} to illustrate the \gls{cip}/\gls{cap} framework: 

\begin{itemize}
	\item Treatment costs: When a patient (either \gls{tp} or \gls{fp}) is included in a home care programme, their \gls{qol} is expected to improve due to increased comfort, reduced hospitalisations, and greater autonomy. Thus, a negative \gls{qol} cost of \(-1.0\) is assigned, which indicates maximum benefits. Simultaneously, healthcare costs decrease due to reduced hospital resource usage, represented as \(-0.5\) in this example. For patients who remain in standard care (\gls{tn}, \gls{fn}), both costs are set to 0, as this reflects the default baseline. 
	\item Errors incur additional burden. In our study, a patient incorrectly classified as at high risk of mortality (\gls{fp}) will receive home care but eventually be rehospitalised, potentially delaying necessary treatment. This results in a moderate \gls{qol} penalty of \(0.5\) and healthcare cost of \(0.25\). In the case of \gls{fn}, a high-risk patient is not assigned to home care, potentially leading to a poor end-of-life experience and avoidable strain on the healthcare system. Hence, this is assigned the highest penalty: \gls{qol} cost of \(1.0\) and healthcare cost of \(1.0\). Correct predictions (\gls{tp}, \gls{tn}) incur no error cost. 
\end{itemize}

Given this baseline definition, the full cost matrix (determined by the consensus of clinicians involved in this project) is summarised in Table~\ref{tab:cost_matrices}. 

\begin{table}[h]
\centering
\caption{\textbf{Assigned quality of life and healthcare costs per outcome scenario}}
\label{tab:cost_matrices}
\footnotesize{
\begin{tabular}{@{}ll
                S[table-format=1.1, table-number-alignment=right]  
                S[table-format=1.2, table-number-alignment=right]@{} 
               }
\toprule
\textbf{Type} & \textbf{Scenario} & {\textbf{\gls{qol} cost}} & {\textbf{Healthcare cost}} \\
\midrule
Treatment & \gls{tp} & -1.0 & -0.5 \\
Treatment & \gls{fp} & -1.0 & -0.5 \\
Treatment & \gls{tn} &  0.0 &  0.0 \\
Treatment & \gls{fn} &  0.0 &  0.0 \\
Error     & \gls{tp} &  0.0 &  0.0 \\
Error     & \gls{fp} &  0.5 &  0.25 \\
Error     & \gls{tn} &  0.0 &  0.0 \\
Error     & \gls{fn} &  1.0 &  1.0 \\
\bottomrule
\multicolumn{4}{@{}p{7cm}@{}}{FN=\textit{\glsentrylong{fn}}. FP=\textit{\glsentrylong{fp}}. QoL=\textit{\glsentrylong{qol}}. TP=\textit{\glsentrylong{tp}}. TN=\textit{\glsentrylong{tn}}.}
\end{tabular}
}
\end{table}

With the definitions of the cost dimensions in place, we compute expected costs at the population level by combining the model’s prediction outputs (\gls{tp}, \gls{fp}, \gls{tn}, \gls{fn}) across a range of decision thresholds from 0 to 1. The \gls{cip} cost curves are then calculated according to Supplementary Section~\hyperref[ssec:cip_implementation]{CIP implementation details}, which aggregates treatment and error costs across the \gls{qol} and healthcare dimensions. 
 
To visualise these components, we introduce a baseline, referred to as the \textit{zero cost curve}, that separates costs (plotted above the baseline) from benefits (plotted below the baseline). Each cost is presented as distinct, colour-coded stacked areas above and below this reference line, respectively, providing an overview of cost composition. 
The upper boundary of the stacked areas (the silhouette of the curve) represents the net effect, calculated as total cost minus total benefit at each threshold. 

\subsubsection[\gls{cap} Step 3]{Step 3: Patient-level cost-benefit analysis using four \gls{llm} agents} \label{sec:step3}
In this step, the system generates a structured clinical cost-benefit analysis tailored to an individual patient.
The core prompt components are detailed in Table~\ref{tab:quadrants}. Detailed prompts are provided in Supplementary Listing~\hyperref[slist:agents_one]{S1}-~\hyperref[slist:agents_sum]{S5}. We used the state-of-the-art \gls{llm} model  gpt-4.1-2025-04-14 to implement the agent responses for the prompt queries. 

\begin{table*}[tbh!]
\centering
\caption{\textbf{Structure of prompts and input dependencies used to construct the four \gls{cap} decision support agents} \newline Each column (I–IV) represents a distinct decision support agent. Rows specify the contextual inputs and response dependencies required for each agent.}
\label{tab:quadrants}
\footnotesize{
\begin{tabular}{|c|p{7cm}|c|c|c|c|}
\toprule
 \textbf{Type} &\textbf{Prompt} & \textbf{I}& \textbf{II} & \textbf{III}& \textbf{IV}\\
\midrule
Context&Patient clinical profile&   \cmark&\cmark  &\cmark& \cmark\\
Context&Classifier description&   \cmark& && \\
Context&Classifier decision threshold $r$  & \cmark& && \\
Context&Classifier performance summary near $r$  &\cmark & && \\
Context&Predicted risk score $s$  & \cmark& \cmark&& \\
Context&Classifier performance summary near $s$  &\cmark & && \\
Context&\gls{cip} cost description  & &\cmark && \\
Context&\gls{cip} cost coefficients  & &\cmark && \\
Context&Composition of \gls{cip} cost curves near $s$&   & \cmark&& \\
\midrule
Context&Response from I&    &  \cmark&  \cmark & \\
Context&Response from II&   &  & &\cmark  \\
\midrule
Query& Classification risk analysis&  \cmark & && \\
Query& Clinical cost-benefit analysis&   & \cmark&& \\
Query& Classification risk mitigation&   & &\cmark& \\
Query& Intervention risk prediction and intervention&   & && \cmark\\
\bottomrule
\multicolumn{6}{@{}p{12cm}@{}}{CAP=\textit{\glsentrylong{cap}}. CIP=\textit{\glsentrylong{cip}}. I=How certain is the risk prediction? II=How do I interpret the \gls{cip} cost curves? III=How can prediction uncertainty be reduced? IV=How can future risks be mitigated?}
\end{tabular}
}
\end{table*}

\subsection{Ethics} 
Ethical approval of study was granted by the Swedish Ethical Review Authority, through the Ethics Committee of the Umeå University (EPN Reference: DNR 2021-02786). 

\section{Results} \label{sec:results}
\subsection{Patient baseline characteristics}
The cohort had a median age of 79 years, 45$\%$ women and median hospital-admittance 6.1 days at baseline, with a 1-year mortality of 22$\%$. Patients who died were older, more often female and with longer hospital stays (85 years, 48$\%$ women, 7.9 days, as compared with 78 years, 44$\%$ women and 5.8 days in survivors) as presented in Table~\ref{tab:baseline}. Statistical analysis was performed as in 
Supplementary Section~\hyperref[ssec:stats_testing]{Statistical testing}.

\begin{table*}[tbh!]
\centering
\caption{\textbf{Baseline characteristics of heart failure cohort} - Table continues next page.}
\label{tab:baseline}
\footnotesize{
\begin{tabular}{@{}lllll@{}}
\toprule
\textbf{Variable} & \textbf{Total} & \textbf{Survived} & \textbf{Deceased within 1 year} & $p$\textbf{-value} \\ \midrule
Number of patients & 30021 & 23499 & 6522 & - \\
\textbf{Age and sex} & & & & \\
Age (years), median (IQR) & 79 (71, 86) & 78 (69, 85) & 85 (78, 90) & \textless{}0.001 \\
\textit{Sex} & & & & \\
Women, n (\%) & 13572 (45) & 10432 (44) & 3140 (48) & \textless{}0.001 \\
BMI (kg/m$^2$), median (IQR) & 26.5 (23.6, 30.4) & 26.8 (23.9, 30.7) & 24.7 (21.8, 28.6) & \textless{}0.001 \\
\textbf{Visit information} & & & & \\
ICU stay, n (\%) & 635 (2.1) & 523 (2.2) & 112 (1.7) & 0.01 \\
ICU days, median (IQR) & 5.3 (2.1, 12.0) & 5.1 (2.1, 10.4) & 7.0 (2.1, 15.9) & 0.04 \\
In-hospital days, median (IQR) & 6.1 (3.1, 10.9) & 5.8 (3.0, 10.0) & 7.9 (4.2, 13.7) & \textless{}0.001 \\
\textbf{Comorbidities, n (\%)} & & & & \\
CCI, median (IQR) & 3 (2, 4) & 2 (1, 4) & 4 (2, 6) & \textless{}0.001 \\
Main heart failure diagnosis & 10993 (37) & 8565 (36) & 2428 (37) & 0.25 \\
Alcohol abuse & 895 (3.0) & 721 (3.1) & 174 (2.7) & 0.1 \\
Aortic aneurysm & 1022 (3.4) & 795 (3.4) & 227 (3.5) & 0.73 \\
Aortic stenosis & 2730 (9.1) & 1955 (8.3) & 775 (12) & \textless{}0.001 \\
Asthma & 2808 (9.3) & 2244 (9.6) & 564 (8.6) & 0.03 \\
Atrial fibrillation & 15485 (52) & 11982 (51) & 3503 (54) & \textless{}0.001 \\
Cancer & 6352 (21) & 4349 (19) & 2003 (31) & \textless{}0.001 \\
Cardiomyopathy & 2346 (7.8) & 2156 (9.2) & 190 (2.9) & \textless{}0.001 \\
Chronic coronary syndrome & 5955 (20) & 4546 (19) & 1409 (22) & \textless{}0.001 \\
Chronic kidney disease & 4515 (15) & 3007 (13) & 1508 (23) & \textless{}0.001 \\
COPD & 4170 (14) & 3016 (13) & 1154 (18) & \textless{}0.001 \\
Diabetes type 1 & 1097 (3.6) & 861 (3.7) & 236 (3.6) & 0.89 \\
Diabetes type 2 & 7598 (25) & 5833 (25) & 1765 (27) & \textless{}0.001 \\
Dyslipidemia & 9666 (32) & 7672 (33) & 1994 (31) & \textless{}0.001 \\
Gonarthrosis & 2505 (8.3) & 1965 (8.4) & 540 (8.3) & 0.85 \\
Heart failure ICD-10 code I50 & 28174 (94) & 21870 (93) & 6304 (97) & \textless{}0.001 \\
Hypertension & 21130 (70) & 16213 (69) & 4917 (75) & \textless{}0.001 \\
Ischemic stroke & 2128 (7.1) & 1544 (6.6) & 584 (8.9) & \textless{}0.001 \\
Leg fracture & 2038 (6.8) & 1400 (6.0) & 638 (9.8) & \textless{}0.001 \\
Myocardial infarction & 4434 (15) & 3572 (15) & 862 (13) & \textless{}0.001 \\
Obesity & 3465 (12) & 2938 (12) & 527 (8.1) & \textless{}0.001 \\
Pulmonary embolism & 1140 (3.8) & 850 (3.6) & 290 (4.5) & \textless{}0.001 \\
Substance abuse & 1938 (6.5) & 1573 (6.7) & 365 (5.6) & \textless{}0.001 \\
Valvular disease & 2873 (9.6) & 2271 (9.7) & 602 (9.2) & 0.3 \\
\textbf{Previous surgery or interventions, n (\%)} & & & & \\
Coronary angioplasty graft & 2371 (7.9) & 1943 (8.3) & 428 (6.6) & \textless{}0.001 \\
Coronary artery bypass grafting & 1798 (6.0) & 1415 (6.0) & 383 (5.9) & 0.67 \\
Heart transplant & 27 (0.1) & 22 (0.1) & 5 (0.1) & 0.86 \\
Pacemaker or defibrillator & 2539 (8.5) & 1949 (8.3) & 590 (9.0) & 0.06 \\
Valve replacement & 1191 (4.0) & 957 (4.1) & 234 (3.6) & 0.08 \\
 \multicolumn{5}{@{}p{15cm}@{}}{\textbf{Prescribed drugs prior to diagnostic hospitalisation, n (\%)}} \\
ACEi & 4875 (16) & 3977 (17) & 898 (14) & \textless{}0.001 \\
Antiarrhythmics, class III & 789 (2.6) & 707 (3.0) & 82 (1.3) & \textless{}0.001 \\
Angiotensin receptor blockers & 3207 (11) & 2526 (11) & 681 (10) & 0.49 \\
ARNI & 330 (1.1) & 298 (1.3) & 32 (0.5) & \textless{}0.001 \\
Beta blockers & 8947 (30) & 7121 (30) & 1826 (28) & \textless{}0.001 \\
Cardiac glycosides & 1893 (6.3) & 1486 (6.3) & 407 (6.2) & 0.83 \\
Cardiac stimulants & 673 (2.2) & 560 (2.4) & 113 (1.7) & \textless{}0.001 \\
Loop diuretics & 10854 (36) & 8003 (34) & 2851 (44) & \textless{}0.001 \\
Thiazide diuretics & 713 (2.4) & 547 (2.3) & 166 (2.5) & 0.33 \\
Diuretics, exclude thiazide & 483 (1.6) & 319 (1.4) & 164 (2.5) & \textless{}0.001 \\
MRA & 4171 (14) & 3322 (14) & 849 (13) & 0.02 \\
SGLT2 inhibitors & 363 (1.2) & 311 (1.3) & 52 (0.8) & \textless{}0.001 \\ \bottomrule
\multicolumn{5}{@{}p{15cm}@{}}{\scriptsize{ACEi=angiotensin-converting enzyme inhibitors. ARNI=angiotensin receptor neprilysin inhibitors. BMI=body mass index. CCI=Charlson comorbidity index. COPD=chronic obstructive pulmonary disease. GFR=glomerular filtration rate. ICU=intensive care unit. IQR=interquartile range. MRA=mineralocorticoid receptor antagonist. NT-proBNP=N-terminal pro b-type natriuretic peptide. SGLT2=sodium-glucose transport protein 2.}}
\end{tabular}
}
\end{table*}

\setcounter{table}{2} 
\begin{table*}[tbh!]
\centering
\caption{\textbf{Baseline characteristics of heart failure cohort} - continued.}
\label{tab:baseline2}
\footnotesize{
\begin{tabular}{@{}lllll@{}}
\toprule
\textbf{Variable} & \textbf{Total} & \textbf{Survived} & \textbf{Deceased within 1 year} & $p$\textbf{-value} \\ \midrule
\multicolumn{5}{@{}p{15cm}@{}}{\textbf{Vital signs, median (IQR)}}\\
Diastolic blood pressure (mmHg) & 74 (69, 80) & 75 (69, 81) & 72 (67, 78) & \textless{}0.001 \\
Systolic blood pressure (mmHg) & 132 (120, 145) & 132 (121, 145) & 129 (117, 142) & \textless{}0.001 \\
Body temperature (°C) & 36.7 (36.4, 36.9) & 36.7 (36.5, 36.9) & 36.6 (36.4, 36.9) & \textless{}0.001 \\
Pulse rate (beats/min) & 78 (70, 88) & 78 (69, 88) & 80 (72, 89) & \textless{}0.001 \\
Respiratory rate (breaths/min) & 19 (17, 22) & 19 (17, 21) & 20 (18, 23) & \textless{}0.001 \\
Oxygen saturation (\%) & 96 (94, 97) & 96 (94, 97) & 95 (93, 96) & \textless{}0.001 \\
\multicolumn{5}{@{}p{15cm}@{}}{\textbf{Laboratories, median (IQR)}}\\
Albumin (34--45 g/L) & 33 (29, 36) & 33 (30, 36) & 31 (27, 34) & \textless{}0.001 \\
Bilirubin (5--25 µmol/L) & 12 (8, 17) & 12 (8, 17) & 12 (8, 17) & 0.02 \\
Blood urea nitrogen (F/M 8.7--22/9.8--23 mg/dL) & 21 (16, 26) & 20 (16, 25) & 24 (19, 28) & \textless{}0.001 \\
C-reactive protein (0--5 mg/L) & 19 (5, 61) & 16 (5, 58) & 30 (10, 71) & \textless{}0.001 \\
Creatinine (F/M 45--90/60--105 µmol/L) & 88 (72, 107) & 86 (71, 105) & 93 (73, 114) & \textless{}0.001 \\
Ferritin (F/M 5--105/27--400 µg/L) & 99 (44, 192) & 100 (44, 190) & 94 (46, 199) & 0.92 \\
Fasting glucose (4--6.9 mmol/L) & 6.1 (5.4, 7.0) & 6.1 (5.4, 7.0) & 6.2 (5.5, 8.0) & 0.1 \\
Plasma glucose (4--6.3 mmol/L) & 6.8 (6.1, 7.6) & 6.8 (6.1, 7.5) & 6.9 (6.2, 7.6) & \textless{}0.001 \\
Hemoglobin (F/M 117--153/134--170 g/L) & 124 (109, 138) & 125 (111, 139) & 118 (104, 131) & \textless{}0.001 \\
Glycated hemoglobin (31--46 mmol/mol) & 39 (35, 45) & 38 (35, 44) & 41 (36, 49) & \textless{}0.001 \\
NT-proBNP (0--400 ng/L) & 4169 (1870, 8843) & 3634 (1641, 7570) & 6917 (3190, 14400) & \textless{}0.001 \\
Potassium (3.5--4.6 mmol/L) & 4.0 (3.8, 4.3) & 4.0 (3.8, 4.3) & 4.1 (3.8, 4.4) & 0.27 \\
Sodium (136--145 mmol/L) & 139 (137, 141) & 139 (137, 141) & 139 (137, 141) & 0.1 \\
Alanine transaminase (F/M 0.25--0.60/0.25--0.75 µkat/L) & 0.4 (0.3, 0.6) & 0.4 (0.3, 0.6) & 0.3 (0.2, 0.5) & \textless{}0.001 \\
Aspartate transaminase (F/M 0.25--0.75/0.25--1.1 µkat/L) & 0.4 (0.3, 0.6) & 0.4 (0.3, 0.6) & 0.4 (0.2, 0.6) & \textless{}0.001 \\
Troponin I (F/M 0--16/0--35 ng/L) & 14 (6, 25) & 13 (6, 25) & 15 (7, 25) & 0.11 \\
Troponin T (0--14 ng/L) & 15 (11, 18) & 15 (11, 18) & 17 (12, 19) & \textless{}0.001 \\
Uric acid (F/M 155--400/230--480 µmol/L) & 416 (322, 513) & 413 (320, 510) & 424 (336, 524) & 0.31 \\
Estimated GFR (mL/min/1.73m$^2$) & 57 (45, 70) & 59 (47, 71) & 50 (40, 63) & \textless{}0.001 \\
\textbf{Physical mobility} & & & & \\
Physical status, median (IQR) & 0 (0, 1) & 0 (0, 1) & 0 (0, 2) & \textless{}0.001 \\
Pressure wounds, n (\%) & 3606 (12) & 2413 (10) & 1193 (18) & \textless{}0.001 \\ \bottomrule
\multicolumn{5}{@{}p{15cm}@{}}{\scriptsize{ACEi=angiotensin-converting enzyme inhibitors. ARNI=angiotensin receptor neprilysin inhibitors. BMI=body mass index. CCI=Charlson comorbidity index. COPD=chronic obstructive pulmonary disease. GFR=glomerular filtration rate. ICU=intensive care unit. IQR=interquartile range. MRA=mineralocorticoid receptor antagonist. NT-proBNP=N-terminal pro b-type natriuretic peptide. SGLT2=sodium-glucose transport protein 2.}}
\end{tabular}
}
\end{table*}

\subsection{Machine learning performance}
We developed various candidate models for the \gls{ml} classifier, and implementation details are outlined in Supplementary Section~\hyperref[ssec:ml_implementation]{Machine learning implementation details} and Supplementary Table~\hyperref[stab:hp_optimisation]{S2}. Figure~\ref{fig:ml_curve_anaylsis} visualises the \emph{receiver operating characteristic} (ROC) curve, \emph{precision-recall curve} (PRC) and calibration curve for all model candidates. Table~\ref{tab:ml_curve_metrics} reports the bootstrapped estimates of \gls{auroc}, \gls{auprc} and \gls{bs} for a more robust performance assessment. Boosting machines outperformed simpler models and were well calibrated according to the ideal calibration line. Due to the highest \gls{auprc} out of all candidates, \gls{xgb} was chosen as the final model for clinical application. The discriminatory capability of \gls{xgb} resulted in \mbox{$\mathrm{\gls{auroc}}=0.804\;(95\%\;$\text{\glsentryshort{ci}}$\;0.792\text{--}0.816)$} and \mbox{$\mathrm{\gls{auprc}}= 0.529\;(95\%\;\text{\gls{ci}}\;0.502\text{--}0.558)$} with a calibration of \mbox{$\mathrm{\gls{bs}}=0.135\;(95\%\;\text{\gls{ci}}\;0.130\text{--}0.140)$}. The decision threshold \mbox{$t_\mathrm{d}=0.25$} was identified at the maximal F1 score \mbox{$\mathrm{F}1_{\max}=0.543$} (Supplementary Fig.~\hyperref[sfig:f1_comparison]{S2}). 

\begin{figure*}[!h]
    \centering
    \includegraphics[width=6.5in]{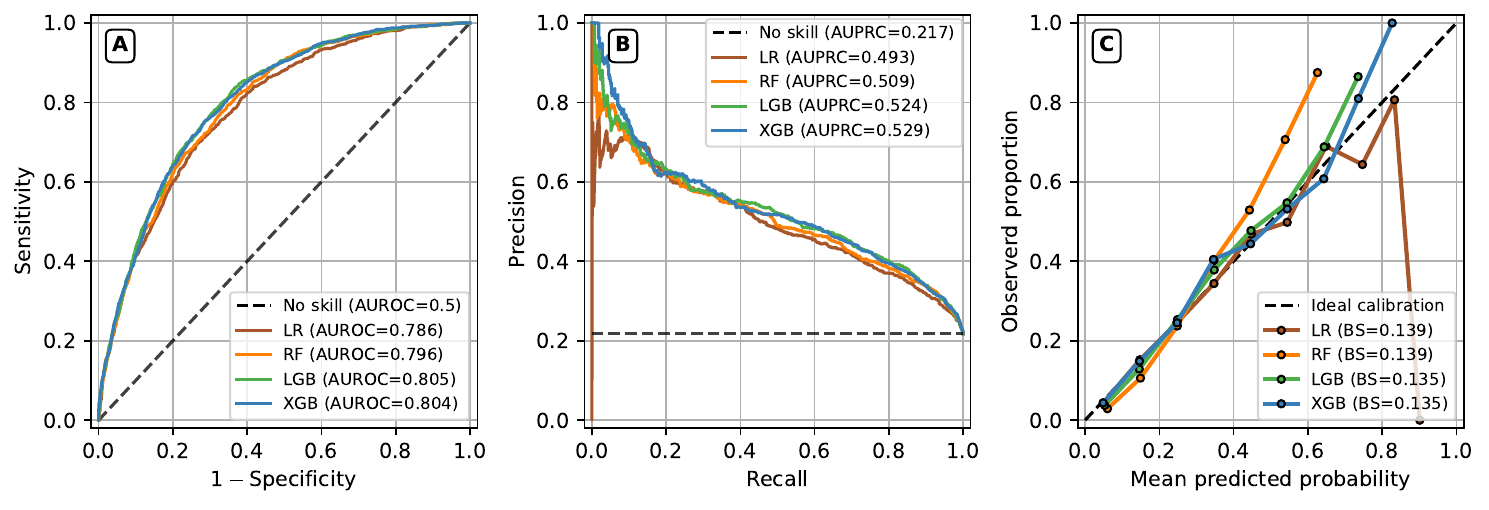}
    \caption{\textbf{Visual comparison of discriminative and calibration performance of model candidates} \newline Receiver operating characteristic curve (\textbf{A}), precision-recall curve (\textbf{B}) and calibration curve (\textbf{C}) demonstrate highest predictive performance for gradient boosting machines based on the test set. AUPRC=\textit{\glsentrylong{auprc}}. AUROC=\textit{\glsentrylong{auroc}}. BS=\text{\glsentrylong{bs}}. LGB=\textit{\glsentrylong{lgb}}. LR=\textit{\glsentrylong{lr}}. RF=\textit{\glsentrylong{rf}}. XGB=\textit{\glsentrylong{xgb}}.}
    \label{fig:ml_curve_anaylsis}
\end{figure*}

\begin{table}[tbh!]
\centering
\caption{\textbf{Bootstrapped performance comparison of model candidates} \newline \gls{auroc}, \gls{auprc} and \gls{bs} are reported with 95$\%$ \glspl{ci}. Best performances are highlighted in bold.}
\label{tab:ml_curve_metrics}
\footnotesize{
\begin{tabular}{@{}llll@{}}
\toprule
\textbf{Model} & \textbf{\gls{auroc}} & \textbf{\gls{auprc}} & \textbf{\gls{bs}} \\ \midrule
LR & 0.786 & 0.494 & 0.139 \\
& (0.772-0.799) & (0.466-0.524) & (0.134-0.144) \\
RF & 0.797 & 0.510 & 0.139  \\
 & (0.784-0.809) & (0.481-0.539) & (0.135-0.144) \\
LGB & \textbf{0.806} & 0.525  & \textbf{0.135}  \\
 &  (0.793-0.818) &  (0.495-0.554) &  (0.130-0.140) \\
XGB & 0.804  & \textbf{0.529}  & \textbf{0.135} \\
& (0.792-0.816) & (0.502-0.558) & (0.130-0.140) \\
\bottomrule
\multicolumn{4}{@{}p{7cm}@{}}{AUPRC=\textit{\glsentrylong{auprc}}. AUROC=\textit{\glsentrylong{auroc}}. BS=\text{\glsentrylong{bs}}. LGB=\textit{\glsentrylong{lgb}}. LR=\textit{\glsentrylong{lr}}. RF=\textit{\glsentrylong{rf}}. XGB=\textit{\glsentrylong{xgb}}.}
\end{tabular}
}
\end{table}

\subsection{\gls{cip} cost curves}
Figure~\ref{fig:cip} presents the \gls{cip} cost curves at different thresholds with a shaded (coloured yellow) risk band, illustrating individual risk, here exemplified for synthetic patient 1. The detailed synthesis process is described in Supplementary Section~\hyperref[ssec:sync_details]{Synthetic patient generation for LLM prompts}. The results from the \gls{cip} visualisations provide insights into the clinical impact at the patient-specific predicted risk in comparison to the decision threshold. Moreover, the risk band highlights the relative change in competing costs. 

\begin{figure}[!h]
    \centering
    \includegraphics[width=0.5\columnwidth]{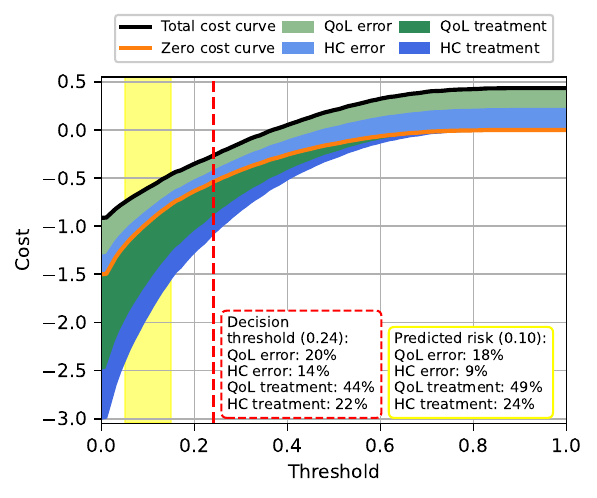}
    \caption{\textbf{\gls{cip} cost curves visualises different cost contributions} \newline \gls{cip} cost curves combine the cost curves for prediction error and treatment costs at varying decision thresholds taking different cost dimensions, namely patient's \gls{qol} and healthcare system, into account. At the population-level, the stacked cost contributions highlight the clinical impact between potentially competing factors. At the patient-level, \gls{cip} visualises the cost-benefit within a risk band (yellow shade) of the patient-specific risk prediction in relation to the decision threshold. CIP=\textit{\glsentrylong{cip}}. HC=health care. QoL=\textit{\glsentrylong{qol}}.}
    \label{fig:cip}
\end{figure}

\subsection{Evaluation by clinicians}
The outcome was evaluated by three clinicians (one clinician for development review and two for user review as in Supplementary Section~\hyperref[ssec:agent_evaluation]{LLM-agent evaluation details}). The review is conducted on 10 synthetic patients (Supplementary Table~\hyperref[stab:synthetic_patients]{S5}). The clinical development review focused on two key aspects: reliability and clinical accuracy, and it included an inventory of common themes of insatisfactory quality in its outputs with respect to the specific \gls{llm}-agents. A prototype of the \gls{cap} framework used for clinical evaluation can be found in Supplementary Fig.~\hyperref[sfig:capboard_patient1]{S4}. 

The evaluation revealed clear differences in the perceived reliability and accuracy of the four agent outputs (Likert scale, 1–5). Agent I ("How certain is the risk prediction?") received the highest ratings, with a reliability score of 4.70 and accuracy score of 4.20 across all patients. While generally well-received, the reviewer noted occasional inappropriate terminology and the inclusion of unsupported advice. Agent II ("How do I interpret the \gls{cip} cost curves?") was rated as reliable (4.00), but accuracy was lower (2.10), due to speculative statements regarding \gls{qol} and healthcare cost consequences. Agent III ("How can prediction uncertainty be reduced?") achieved moderate ratings (reliability 3.60, accuracy 2.30), with qualitative feedback indicating that the advice was sometimes overconfident and partially unsubstantiated. Agent IV ("How can future risks be mitigated?") received the lowest scores (reliability 2.60, accuracy 1.70), with the reviewer describing the guidance as unrealistic and frequently ungrounded. 

We categorised the identified themes into 11 categories, which were annotated during the development review. The five most common issues were: (1) incorrect or overreaching medical terminology; (2) overly confident and imperative advice (e.g. use of "shall", "must"); (3) unrealistic, unfeasible, or idealistic recommendations; (4) unasked-for or unsolicited advice; and (5) overbearing or overly prescriptive advice concerning minor details. 

The clinical user review was conducted across the same 10 patient cases, with clinicians providing both Likert-scale ratings (1–5) and qualitative comments for each of the \gls{cap} components. For risk estimation, the users reported high confidence when using the classifier’s risk estimate in combination with the tool, with consistent feedback that the tool enhanced their risk assessment process. The generated risk explanations were generally seen as clear and useful, though in some cases they added little beyond what the risk score already conveyed. The \gls{cip} cost curves received mixed feedback. The users found the concept useful for illustrating trade-offs. The tool’s uncertainty handling was appreciated, particularly its suggestions for reducing uncertainty, though these were sometimes too generic or incomplete for very complex cases. Importantly, the tool did not create a false sense of reassurance; the user consistently noted it maintained appropriate caution. 

\section{Discussion}

Our work focuses on the development of \emph{interpretable} predictive models for 1-year all-cause mortality in heart failure, structured around three central questions. Q1: Why does the model make a positive or negative prediction? Q2: What are the consequences of acting on the model’s predictions in clinical care? Q3: How sensitive and robust are these predictions when used in practice? 

\subsection{European Union \gls{ai} legal requirements}
Importantly, the European Union \gls{ai} act poses legally binding requirements on transparency, safety, accountability and ethics for development of \gls{ai} to be applied in health care (Artificial intelligence act \cite{eu-2024/1689}). In particular, the EU \gls{ai} Act requires clear definition, assessment, and governance of risks arising from the output of \gls{ai} models. \gls{cip} addresses this by categorising risks into patient- and healthcare system–related dimensions, thereby aligning with the Act’s requirements. Each cost dimension reflects a distinct risk, presented within a structured framework where risks are evaluated based on both technical performance and real-world clinical and economic impact. Notably, the risks here should be read as an relevant example of important considerations in heart failure but may be tailored to the user´s needs. Moreover, the evaluation of these costs facilitates effective and transparent communications of risks, as stakeholders can clearly see the trade-offs between different decisions. Thus, the cost evaluation framework will aid in initial risk assessment but also enable the long-term safety, effectiveness, and accountability of the \gls{ai} system. 

\subsection{Clinical feedback on \gls{llm}-agents}
The cost-aware decision guidance, communicated by the \gls{llm}-agents, was rated positively and was helpful for assessing future risk. However, the integration of patient-level and population-level information had a steep learning curve for the evaluating clinicians. The generated explanations were generally well-received. The users felt the explanations captured key decision factors and expressed confidence about applying them in practice. There was little concern about content falling outside established guidelines or evidence-based practice, though in complex cases, explanations included unnecessary details or became overly lengthy. 

The clinical user review shows that the system was generally well-received for its ability to support risk estimation and explanation, suggesting that the \gls{llm} agents are most effective (accurate and helpful) in descriptive and explanatory tasks. However, speculative or forward-looking guidance was unsatisfactory and requires further development. 
Overall, the clinical user review results suggest the system’s strengths lie in supporting risk communication and generating structured clinical explanations. 

\subsection{Potential applications for policymakers and healthcare financiers}
Note that although this paper addresses clinicians as stakeholders, it may also be relevant to hospital administrators and policymakers, and can be adjusted according to the selection of included costs. For example, whereas the correctly classified groups, \gls{tp} and \gls{tn}, are clinically uncomplicated, the \gls{fn} illustrate delivery of care that does not contribute to patient value although a costly choice in our model. For comparable conditions, the framework may be useful for policymakers when planning the distribution of hospital beds versus home care. 

\subsection{Related work}
Previous research has successfully demonstrated the application of \gls{ml} models for prediction of all-cause mortality within 1 year for hospitalised heart failure patients \cite{
adler_improving_2020, angraal_machine_2020,
tohyama_machine_2021, 
takahama_clinical_2023,
tian_machine_2023, ketabi_predicting_2024}. \gls{ml} models outperformed conventional heart failure risk score models \cite{adler_improving_2020,
tohyama_machine_2021, takahama_clinical_2023}.  
Among the various types of \gls{ai} models, \textit{supervised classification models} are generally the most accurate and reliable for clinical prediction tasks. These \gls{ml} models follow well-established methods for development, validation, and evaluation, which favour trust in their results in clinical settings. 
Models of increasing complexity, such as neural networks \cite{kwon_artificial_2019, wang_feature_2020, li_deep_2023} and transformer-based architectures \cite{pang2021cehr, antikainen_transformers_2023}, show potential for further gains but typically require larger datasets, may be less robust (high variance), and present greater interpretability challenges. 
Given the importance of interpretability (Q1) complex model architectures were not prioritised. 

Despite advances in model development, the clinical utility of \gls{ml} models remains limited, and the evaluation of downstream effects of decisions seldom addressed or evaluated (Q2). In clinical practice, risk thresholds for management decisions are rarely well defined but rest on clinical judgement, and the consequences of acting on predictions are often unclear. 

Frameworks such as the \gls{mct} \cite{greiner_principles_2000} and \gls{dca} \cite{vickers_decision_2006} aim to bridge this gap. \gls{mct} quantifies optimal thresholds by balancing sensitivity, specificity, prevalence, and predefined costs of \gls{fp} and \gls{fn}. However, \gls{mct} tightly couples model behaviour to fixed cost terms defined during development, limiting flexibility.  
\gls{dca} visualises net benefit across thresholds, comparing model-guided decisions to treat-none and treat-all strategies \cite{li_prediction_2021, li_predicting_2022, chen_machine_2023, tian_machine_2023}, yet does not break down the cost structure of individual predictions. In contrast, the \gls{cip} cost curve provides patient-level cost composition and by illustrating how sensitive these costs are to threshold changes (Q3). \glspl{dca} constructed for all model candidates can be found in Supplementary Fig.~~\hyperref[sfig:dca_all]{S3}. 
 
To enhance interpretability, \glspl{llm} have shown promise in clinical decision support \cite{shool2025systematic, vrdoljak2025review, zhang2025revolutionizing, denecke2024potential}, including summarising medical records and assisting with diagnostic reasoning. However, their integration into clinical workflows remains at an early stage. Recent studies have highlighted limitations \cite{LI20251, liu-etal-2024-large}, such as inflexible reasoning, overconfidence in outputs, and susceptibility to errors in complex clinical scenarios. Moreover, \glspl{llm} are not inherently reliable for speculative tasks or generating outcomes beyond grounded evidence, as they are prone to hallucinations \cite{shool2025systematic, vrdoljak2025review, zhang2025revolutionizing, denecke2024potential}. Evaluations indicate that \glspl{llm} may underperform compared with human clinicians, particularly when processing unstructured or nuanced patient data. Findings from our exploratory study are consistent with these observations. Nonetheless, \gls{llm}-based approaches hold considerable potential, and further development with rigorous evaluation is warranted. 

\subsection{Strengths and limitations}
Our model is built on a large, comprehensive, and representative regional cohort, ensuring reliable risk mapping, \gls{cip}/\gls{cap}, and reflecting real-world expectations for healthcare services. It can also simulate outcomes with cost considerations to support threshold-setting in healthcare planning. However, data drift over time necessitates regular updates and validation on contemporary cohorts in the evolving and changing clinical environments. While \gls{llm}–based explanations were generally well received, their performance in speculative reasoning and complex patient scenarios remains limited. Further refinement is needed to ensure their safe and reliable clinical use. 

\section{Conclusions}
With \gls{cap} we propose a three-stage framework:

Firstly, we showed that \gls{ml} models can effectively predict all-cause mortality within 1 year in a heart failure cohort with a first in-hospital diagnosis, with \gls{xgb} identified as the best model based on the highest \gls{auprc}. 

Secondly, on population-level we theoretically outlined cost matrices including treatment and prediction error costs for two relevant cost dimensions, related to the patient's \gls{qol} and the healthcare provider. We then conceptualised \gls{cip} to reflect the cost contributions associated with the predictions of the \gls{xgb} classifier across various risk thresholds expanding traditional \gls{ml} metrics by accounting for the clinical impact on patient's \gls{qol} and healthcare costs. 

Thirdly, we integrated \gls{llm} agents into the \gls{cap} framework to communicate four different aspects of output interpretation, for patient-level decision support. 

\section{\normalsize{Supplementary information}}
Supplementary information is available in the appendix.

\tabsup{stab:data_collection}{1}
\tabsup{stab:hp_optimisation}{2}
\tabsup{stab:ml_shap}{3}
\tabsup{stab:best_clf_set_metrics}{4}
\tabsup{stab:synthetic_patients}{5}
\tabsup{stab:cap_questionnaire}{6}
\tabsup{stab:cap_components}{7}
\tabsup{stab:cap_category_feedback}{8}
\figsup{sfig:shap_summary}{1}
\figsup{sfig:f1_comparison}{2}
\figsup{sfig:dca_all}{3}
\figsup{sfig:capboard_patient1}{4}
\figsup{slist:agents_one}{1}
\figsup{slist:agents_sum}{5}
\secsup{ssec:ml_metrics}{}
\secsup{ssec:ml_implementation}{}
\secsup{ssec:cip_implementation}{}
\secsup{ssec:stats_testing}{}
\secsup{ssec:sync_details}{}
\secsup{ssec:agent_evaluation}

\section{\normalsize{Contributors}}
YY: Conceptualization, Data curation, Formal analysis, Funding acquisition, Investigation, Methodology, Project administration, Resources, Software, Supervision, Validation, Visualization, Writing – original draft, Writing – critical review, commentary \& editing.
FD: Conceptualization, Data curation, Formal analysis, Investigation, Methodology, Project administration, Software, Visualization, Writing – original draft, Writing – critical review, commentary \& editing.
MA: Conceptualization, Data curation, Formal analysis, Methodology, Software, Writing – critical review, commentary \& editing.
HS: Conceptualization, Funding acquisition, Methodology, Project administration, Resources, Supervision, Validation, Writing – original draft, Writing – critical review, commentary \& editing.
CEL: Methodology, Writing – critical review, commentary \& editing.
ML: Methodology, Writing – critical review, commentary \& editing.
AR: Resources, Writing – critical review, commentary \& editing.

\section{\normalsize{Declaration of interests}}
All authors declare no conflict of interest. 

\section{\normalsize{Data sharing}}
The data and source code are not publicly available but will be made available upon reasonable request after internal review. 

\section{\normalsize{Acknowledgments}}
This work was supported by grants from the Swedish state under an agreement concerning research and education of doctors (ALFGBG-991470 [to H.S.], ALFGBG-971608 [to M.L.]), the Swedish Research Council (2023-06421 [to H.S.], 2023-02144 [to A.R.]), the Swedish Heart and Lung Foundation (2024-0678 [to A.R.]), and Vinnova Advanced Digitalization (DNR 2024-01446 [to Y.Y. and H.S.]).  

\bibliographystyle{vancouver}
\bibliography{references}



\section*{Supplementary material}

# Supplementary information

**Table S1:** Data collection with detailed information, consisting of outcome for categorical and units for numerical features along with missing percentage, and codes for clinical variables. ACEi=angiotensin-converting enzyme inhibitors. ARNI=angiotensin receptor neprilysin inhibitors. BMI=body mass index. CCI=Charlson comorbidity index. COPD=chronic obstructive pulmonary disease. GFR=glomerular filtration rate. ICU=intensive care unit. MRA=mineralocorticoid receptor antagonist. NT-proBNP=N-terminal pro b-type natriuretic peptide. SGLT2=sodium-glucose transport protein 2. Table continues next page.

| Variable | Outcome/Unit | Details (e.g., codes for extraction) | Missing ratio in % |
|---|---|---|---|
| **Demographics** | | | |
| Age | years | | 0 |
| Sex | 0: male, 1: female | | 0 |
| BMI | kg/m$^2$ | | 54.69 |
| **Visit information** | | | |
| ICU stay | 0: no, 1: yes | | 0 |
| ICU days | days | | 97.88 |
| In-hospital days | days | | 0 |
| **Comorbidities** | | | |
| CCI | 1 | CCI was caluclated using the comorbidipy package version 0.5.0 with mapping variant charlson and weightning variant se. | 0 |
| Main heart failure diagnosis | 0: no, 1: yes | Qualifying heart failure diagnosis at primary position | 0 |
| Alcohol abuse | 0: no, 1: yes | F10 | 0 |
| Aortic aneurysm | 0: no, 1: yes | I71 | 0 |
| Aortic stenosis | 0: no, 1: yes | I350, I352 | 0 |
| Asthma | 0: no, 1: yes | J45 | 0 |
| Atrial fibrillation | 0: no, 1: yes | I48 | 0 |
| Cancer | 0: no, 1: yes | C | 0 |
| Cardiomyopathy | 0: no, 1: yes | I42, I43 | 0 |
| Chronic coronary syndrome | 0: no, 1: yes | I259 | 0 |
| Chronic kidney disease | 0: no, 1: yes | N18 | 0 |
| COPD | 0: no, 1: yes | J44 | 0 |
| Diabetes type 1 | 0: no, 1: yes | E10 | 0 |
| Diabetes type 2 | 0: no, 1: yes | E11 | 0 |
| Dyslipidemia | 0: no, 1: yes | E78 | 0 |
| Gonarthrosis | 0: no, 1: yes | M179 | 0 |
| Heart failure | 0: no, 1: yes | I50 | 0 |
| Hypertension | 0: no, 1: yes | I10, I11, I12, I13, I15 | 0 |
| Ischemic stroke | 0: no, 1: yes | I63 | 0 |
| Leg fracture | 0: no, 1: yes | S72, S82, S92 | 0 |
| Myocardial infarction | 0: no, 1: yes | I21 | 0 |
| Obesity | 0: no, 1: yes | E66 | 0 |
| Pulmonary embolism | 0: no, 1: yes | I26 | 0 |
| Substance abuse | 0: no, 1: yes | F11, F12, F13, F14, F15, F16, F17, F18, F19 | 0 |
| Valvular disease | 0: no, 1: yes | I05, I06, I07, I08, I09, I33, I34, I351, I36, I37, I38, I39 | 0 |
| **Previous surgery or interventions** | | | |
| Coronary angioplasty graft | 0: no, 1: yes | Z955 | 0 |
| Coronary artery bypass grafting | 0: no, 1: yes | Z951 | 0 |
| Heart transplant | 0: no, 1: yes | Z941 | 0 |
| Pacemaker or defibrillator | 0: no, 1: yes | Z950 | 0 |
| Valve replacement | 0: no, 1: yes | Z952, Z953, Z954 | 0 |
| **Prescribed drugs prior to diagnostic hospitalisation** | | | |
| ACEi | 0: no, 1: yes | C09AA01, C09AA02, C09AA03, C09AA05, C09AA10 | 0 |
| Antiarrhythmics, class III | 0: no, 1: yes | C01BD | 0 |
| Angiotensin receptor blockers | 0: no, 1: yes | C09CA01, C09CA03, C09CA06 | 0 |
| ARNI | 0: no, 1: yes | C09DX04 | 0 |
| Beta blockers | 0: no, 1: yes | C07AB02, C07AB07, C07AG02, C07AB12 | 0 |
| Cardiac glycosides | 0: no, 1: yes | C01A | 0 |
| Cardiac stimulants | 0: no, 1: yes | C01C | 0 |
| Loop diuretics | 0: no, 1: yes | C03C | 0 |
| Thiazide diuretics | 0: no, 1: yes | C03A | 0 |
| Diuretics, exclude thiazide | 0: no, 1: yes | C03B | 0 |
| MRA | 0: no, 1: yes | C03DA | 0 |

| Variable | Outcome/Unit | Details (e.g., codes for extraction) | Missing ratio in % |
|---|---|---|---|
| SGLT2 inhibitors | 0: no, 1: yes | A10BK | 0 |
| **Vital signs** | | | |
| Diastolic blood pressure | mmHg | | 5.78 |
| Systolic blood pressure | mmHg | | 5.11 |
| Body temperature | °C | | 6.49 |
| Pulse rate | beats/min | | 5.73 |
| Respiratory rate | breaths/min | | 15.41 |
| Oxygen saturation | % | | 6.04 |
| **Laboratories** | | | |
| Albumin | g/L | | 27.81 |
| Bilirubin | µmol/L | | 40.82 |
| Blood urea nitrogen | mg/dL | | 74.06 |
| C-reactive protein | mg/L | | 8.90 |
| Creatinine | µmol/L | | 11.95 |
| Ferritin | µg/L | | 87.42 |
| Fasting glucose | mmol/L | | 95.93 |
| Plasma glucose | mmol/L | | 16.39 |
| Hemoglobin | g/L | | 4.17 |
| Glycated hemoglobin | mmol/mol | | 86.81 |
| NT-proBNP | ng/L | | 41.67 |
| Potassium | mmol/L | | 4.44 |
| Sodium | mmol/L | | 4.34 |
| Alanine transaminase | µkat/L | | 44.73 |
| Aspartate transaminase | µkat/L | | 45.88 |
| Troponin I | ng/L | | 79.23 |
| Troponin T | ng/L | | 95.62 |
| Uric acid | µmol/L | | 97.02 |
| Estimated GFR | mL/min/1.73m$^2$ | Estimated GFR was calculated using the Lund-Malmö formula LMR18. | 11.95 |
| **Physical mobility** | | | |
| Physical status | 0: walking without aid, 1: walking with aid, 2: wheelchair, 3: bedridden | | 0 |
| Pressure wounds | 0: no, 1: yes | | 0 |

**Clinical consequences and interpretations**

This section details the clinical consequences of our application. The implications of misclassification require careful consideration. A FP prediction of high mortality risk may lead to premature withdrawal of hospital-based treatment. This risk, while serious, can be mitigated through open-ended treatment plans that allow clinicians to override model-based recommendations based on evolving clinical assessments. Importantly, such flexibility must be maintained, as AI-based models are subject to stricter regulatory requirements than traditional heuristic tools, such as clinical risk scores. Conversely, FNs risk depriving patients of appropriate home-based palliative care, leading to unnecessary aggressive treatment and increased healthcare costs. In this context, the structured explanations generated by the LLM agents can support clinicians in identifying when model outputs warrant closer scrutiny, potentially enhancing the transparency and safety of decision-making. Furthermore, explicit modelling of the costs associated with FPs and FNs provides valuable insight for policymakers and healthcare managers concerned with both clinical outcomes and resource allocation.

**Machine learning model metrics**

Developing the ML model involves several technical design choices, including selecting the appropriate model type and tuning its hyperparameters. Model performance is evaluated on a test cohort with real-world electronic health record data, using standard ML metrics such as precision and recall. Precision is defined as the proportion of positive identifications that are actually correct (e.g., the proportion of patients correctly identified as likely to die within a year out of all patients identified as such). Recall is the proportion of actual positives that are correctly identified (e.g., the proportion of patients who died within a year and were correctly identified by the model). We adopted the commonly used ML metric to evaluate and compare the performance of different models: the area under the precision-recall curve with varying thresholds. This metric can be used to automatically select the best overall performing model. To define an operational point of the best classifier for decision-making, the decision threshold is selected as the one that maximises the F1

score across all thresholds, where the F1 score measures the relationship between precision and recall as a single metric calculated as harmonic mean.

**Machine learning implementation details**

To predict mortality within 1 year using ML, patient data was divided into 24,016 samples for model training and 6,005 samples for model evaluation stratified for mortality outcome. Four frequently used ML models were employed: logistic regression (LR), random forest (RF), light gradient boosting machine (LGB) and eXtreme gradient boosting (XGB). ML code and models were implemented in Python 3.10.13 with package versions scikit-learn 1.4.0, lightgbm 4.3.0 and xgboost 2.0.0.

Features with more than 80% missing data were removed (Table S1). To address the remaining missingness, features were imputed by the median. As an alternative to median imputation, k-Nearest Neighbors imputation was applied but yielded similar results and was not considered further due to its higher computational cost. Numerical and ordinal features were normalised to $[0,1]$ and categorical features label encoded.

For all model candidates, hyperparameters were optimised using grid search and stratified 5-fold cross-validation on the training set to ensure equal class distribution in all folds. Models were evaluated by maximising the area under the precision-recall curve to address class imbalance. The optimised hyperparameters for all models are reported in Table S2. The best models were then retrained on the complete training set and evaluated on the unseen test set.

The model's discriminatory capability was assessed using receiver operating characteristic (ROC) curve and precision-recall curve (PRC) with area under ROC curve (AUROC) and area under the precision-recall curve (AUPRC) as threshold-independent metrics to summarise each curve. For the reliability of predicted probabilities, calibration curve and Brier score (BS) were evaluated. The 95% confidence intervals (CI) for AUROC, AUPRC and BS were calculated with bootstrapping (1000 iterations). The best model candidate was identified based on the highest AUPRC over AUROC and the respective decision threshold was determined by the highest F1 score to account for class imbalance in the data.

**CIP implementation details**

The purpose of the CIP curve is to visualise how costs are distributed across decision thresholds. Users need to define the relevant cost terms for each clinical dimension (QoL and healthcare provider cost) and cost type. In particular, we consider two cost types in this work,

- Treatment cost ($C^T$): The inherent cost of delivering a given care pathway (for example, hospital-based care or home-based care), independent of prediction accuracy. This is defined relative to a baseline of the standard treatment (for example, hospital-based care).
- Error cost ($C^E$): The additional cost arising from false positive or false negative predictions. This cost is defined relative to the ground truth label.

Let $p_y(t)$ denote the empirical proportion of patients classified into outcome $y$ at threshold $t$. The expected cost for a given cost type $\tau$ and dimension $k$ at threshold $t$ is defined as:

$$\mathbb{E}(C_k^\tau; t) = \sum_{y \in \{\text{TP,FP,FN,TN}\}} p_y(t) \cdot C_k^\tau(y) \qquad (1)$$

where

- $k \in \{1,2\}$ denotes the cost dimension: $C_1$ for QoL cost and $C_2$ for healthcare system cost;
- $\tau \in \{T, E\}$ denotes the cost type: Treatment ($T$) or Error ($E$);
- $y \in \{\text{TP, FP, FN, TN}\}$ represents the prediction outcome;
- $t \in [0,1]$ is the decision threshold applied to the classifier.

The expected cost varies depending on the decision threshold. By systematically evaluating a range of thresholds, we obtain the CIP curve – a population-level overview of the clinical cost implications expressed through these costs.

**Statistical testing**

To determine statistical significance ($p$-value $\leq 0.05$) between non-events and events (deceased within 1 year), Chi-square test was utilised for categorical features (clinical variables), $t$-test for two independent samples for normally distributed numerical features and Mann-Whitney $U$ rank test for non normally distributed numerical features.

**Table S2:** Hyperparameter optimisation setup using grid search with 5-fold cross-validation. LGB=light gradient boosting machine. LR=logistic regression. RF=random forest. XGB=eXtreme gradient boosting.

| Model | Parameter | Search space | Final value |
|---|---|---|---|
| LR | solver | [liblinear, saga] | saga |
| | penalty | [l1, l2] | l1 |
| | C | [0.1, 1, 10, 100] | 1 |
| RF | max_depth | [4, 8, 12] | 12 |
| | n_estimators | [100, 300, 500] | 500 |
| | min_samples_split | [2, 5, 10] | 10 |
| | min_samples_leaf | [1, 2, 4] | 1 |
| LGB | max_depth | [4, 8, 12] | 12 |
| | n_estimators | [100, 300, 500] | 500 |
| | learning_rate | [0.01, 0.1, 0.3] | 0.01 |
| | subsample | [0.5, 0.8, 1] | 0.5 |
| | colsample_bytree | [0.5, 0.8, 1] | 0.5 |
| | reg_alpha | [0, 1, 5] | 0 |
| | reg_lambda | [0, 1, 5] | 5 |
| | num_leaves | [25, 50] | 50 |
| XGB | max_depth | [4, 8, 12] | 8 |
| | n_estimators | [100, 300, 500] | 500 |
| | learning_rate | [0.01, 0.1, 0.3] | 0.01 |
| | subsample | [0.5, 0.8, 1] | 0.5 |
| | colsample_bytree | [0.5, 0.8, 1] | 0.8 |
| | reg_alpha | [0, 1, 5] | 0 |
| | reg_lambda | [0, 1, 5] | 0 |

**Synthetic patient generation for LLM prompts**

To evaluate the large language model (LLM) agents, synthetic patients were generated to allow for the reproducibility of the prompts used in the cost-aware prediction (CAP) framework. The design of the synthetic patient is limited to 20 features as the synthesis of nearly real-world patients is non-trivial and beyond the scope of this work. Each synthetic patient was reviewed by the clinicians in the project. The 20 features, denoted as the core set $\mathcal{S}_{\text{core}}$, include a demographic triplet (age, sex, body mass index) and 17 non-demographic features with the highest contribution to the decision-making by the best classifier. Independent of the assigned contribution, the demographic triplet was included to contextualise the synthetic patient profiles (e.g., sex-dependent laboratory measurements).

The contribution of each feature was evaluated by the SHapley Additive exPlanation (SHAP) method. SHAP values are commonly used to explain the decision process of an ML model by quantifying the average impact of each individual feature on predictions across all possible feature combinations. Figure S1 summarises the SHAP distribution for the top 20 predictive features using XGB which yields the highest AUPRC performance (best classifier). Meanwhile, Table S3 presents the mean SHAP values for all features. To evaluate the competitiveness of XGB when the real patient data was restricted to the core set $\mathcal{S}_{\text{core}}$, the classifier was retrained on $\mathcal{S}_{\text{core}}$ following the previously established implementation pipeline. The hyperparameter tuning was omitted by reusing the hyperparameter setup optimised for $\mathcal{S}_{\text{complete}}$ (69 features after preprocessing). Figure S2 highlights the comparison of the F1 score across various risk thresholds, demonstrating that the core set $\mathcal{S}_{\text{core}}$ is competitive with $\mathcal{S}_{\text{complete}}$ while resulting in a similar decision threshold. Furthermore, Table S4 indicates only a minor performance drop across AUROC, AUPRC and BS.

Since the performance comparison has verified that the reduced classifier performs nearly equally well, 10 synthetic patients were generated using the features in $\mathcal{S}_{\text{core}}$ to utilise their profiles as prompt input for the LLM agents. The 10 synthetic profiles and their respective risk predictions are outlined in Table S5.

**Table S3:** Feature importance of best model (XGB) calculated by mean SHAP values. ACEi=angiotensin-converting enzyme inhibitors. ARNI=angiotensin receptor neprilysin inhibitors. BMI=body mass index. CCI=Charlson comorbidity index. COPD=chronic obstructive pulmonary disease. GFR=glomerular filtration rate. ICU=intensive care unit. MRA=mineralocorticoid receptor antagonist. NT-proBNP=N-terminal pro b-type natriuretic peptide. SGLT2=sodium-glucose transport protein 2. SHAP=SHapley Additive exPlanations. XGB=eXtreme gradient boosting.

| Variable | SHAP$_{mean}$ | Variable | SHAP$_{mean}$ |
| --- | --- | --- | --- |
| Age | 0.4912 | Substance abuse | 0.0086 |
| CCI | 0.2756 | Troponin I | 0.0072 |
| Oxygen saturation | 0.214 | Chronic kidney disease | 0.0068 |
| Albumin | 0.1795 | Alcohol abuse | 0.0067 |
| Body temperature | 0.1597 | Obesity | 0.0061 |
| Systolic blood pressure | 0.1347 | Myocardial infarction | 0.0061 |
| NT-proBNP | 0.1309 | Antiarrhythmics, class III | 0.0053 |
| Respiratory rate | 0.0905 | Leg fracture | 0.0049 |
| Physical status | 0.0889 | Beta blockers | 0.0046 |
| Pulse rate | 0.0884 | Ischemic stroke | 0.0045 |
| Hemoglobin | 0.0829 | Diabetes type 2 | 0.0044 |
| BMI | 0.0759 | Hypertension | 0.0041 |
| In-hospital days | 0.0694 | Main heart failure diagnosis | 0.004 |
| C-reactive protein | 0.0693 | ACEi | 0.0036 |
| Sodium | 0.0646 | Chronic coronary syndrome | 0.0035 |
| Cancer | 0.0617 | Valve replacement | 0.0019 |
| Estimated GFR | 0.0525 | Pacemaker or defibrillator | 0.0019 |
| Loop diuretics | 0.0498 | MRA | 0.0018 |
| Potassium | 0.0463 | Cardiac glycosides | 0.0015 |
| Blood urea nitrogen | 0.0445 | Angiotensin receptor blockers | 0.0014 |
| Sex | 0.0404 | Aortic aneurysm | 0.0013 |
| Diastolic blood pressure | 0.0387 | Diuretics, exclude thiazide | 0.0013 |
| Pressure wounds | 0.0341 | Coronary artery bypass grafting | 0.0012 |
| Alanine transaminase | 0.0312 | Thiazide diuretics | 0.0011 |
| Creatinine | 0.0284 | Valvular disease | 0.0011 |
| Plasma glucose | 0.0251 | Diabetes type 1 | 0.0011 |
| Aspartate transaminase | 0.024 | ICU stay | 0.0009 |
| Bilirubin | 0.0205 | Heart failure | 0.0009 |
| Aortic stenosis | 0.0179 | Coronary angioplasty graft | 0.0009 |
| Dyslipidemia | 0.0124 | Pulmonary embolism | 0.0005 |
| Cardiomyopathy | 0.0117 | SGLT2 inhibitors | 0.0005 |
| Asthma | 0.0115 | Cardiac stimulants | 0.0003 |
| Atrial fibrillation | 0.0114 | ARNI | 0.0002 |
| Gonarthrosis | 0.0112 | Heart transplant | 0.0 |
| COPD | 0.0112 | | |

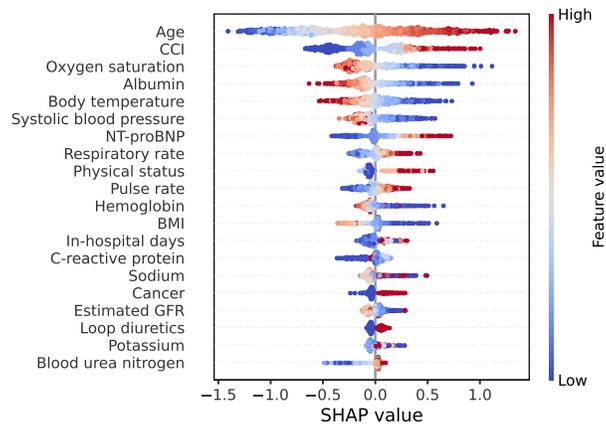

**Figure S1:** Top 20 clinical predictors of best model (XGB) visualised by the SHAP values of each sample and ranked by mean SHAP values. SHAP=SHapley Additive exPlanations. XGB=eXtreme gradient boosting.

5

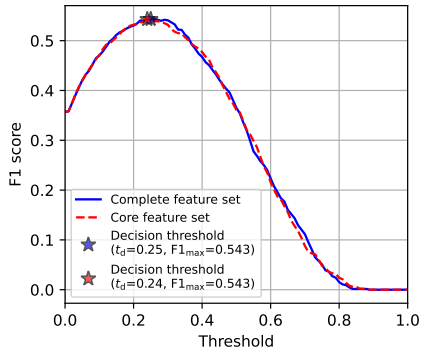

**Table S4:** Performance comparison of the XGB model trained and evaluated either on the complete set of clinical variables $\mathcal{S}_{\text{complete}}$ or the core set of clinical variables $\mathcal{S}_{\text{core}}$. AUROC, AUPRC and BS are reported with 95% CI. AUPRC=area under the precision-recall curve. AUROC=area under the receiver operating characteristic curve. BS=Brier score. XGB=eXtreme gradient boosting.

| Model | Set | AUROC | AUPRC | BS |
|---|---|---|---|---|
| XGB | $\mathcal{S}_{\text{complete}}$ (69 features) | 0.804 (0.792-0.816) | 0.529 (0.502-0.558) | 0.135 (0.130-0.140) |
| XGB | $\mathcal{S}_{\text{core}}$ (20 features) | 0.802 (0.789-0.815) | 0.520 (0.494-0.549) | 0.135 (0.131-0.140) |

**Figure S2:** Performance comparison of F1 score against various thresholds for the best model (XGB) trained on the original high-dimensional feature space $\mathcal{S}_{\text{complete}}$ (69 features) versus a core set $\mathcal{S}_{\text{core}}$ (20 features). Evaluation was performed on the same test set, where each instance was represented by either the complete set or core set of clinical variables. XGB=eXtreme gradient boosting.

**Table S5:** Overview of synthetic heart failure patients. The respective risks are predicted by the best classifier (XGB). Values highlighted in grey simulate missing values that are imputed. Alb=albumin. BMI=body mass index. C=cancer. CCI=Charlson comorbidity index. CRP=C-reactive protein. Days=in-hospital days. eGFR=estimated glomerular filtration rate. Hb=hemoglobin. HD=loop diuretics. Id=patient identifier. K=potassium. Na=sodium. NT-proBNP=N-terminal pro b-type natriuretic peptide. $O_2$=oxygen saturation. PR=pulse rate. PS=physical status. RR=respiratory rate. SBP=systolic blood pressure. Temp=body temperature. XGB=eXtreme gradient boosting.

| Id | Age | Sex | BMI | CCI | $O_2$ | Alb | Temp | SBP | NT-proBNP | RR | PS | PR | Hb | Days | CRP | Na | C | eGFR | HD | K | Risk |
|---|---|---|---|---|---|---|---|---|---|---|---|---|---|---|---|---|---|---|---|---|---|
| 1 | 86 | 0 | 26.5 | 2 | 85 | 37 | 37.2 | 161 | 2424 | 17 | 1 | 75 | 145 | 5.1 | 20 | 140 | 0 | 52 | 0 | 4.1 | 0.010 |
| 2 | 70 | 0 | 26.5 | 1 | 95 | 33 | 36.5 | 134 | 252 | 19 | 1 | 81 | 144 | 2.1 | 20 | 141 | 0 | 55 | 0 | 4.0 | 0.025 |
| 3 | 85 | 0 | 26.5 | 2 | 90 | 33 | 37.0 | 146 | 4142 | 19 | 1 | 82 | 151 | 6.5 | 22 | 140 | 0 | 56 | 0 | 4.2 | 0.208 |
| 4 | 65 | 1 | 26.5 | 7 | 92 | 30 | 38.2 | 154 | 4142 | 17 | 0 | 75 | 170 | 5.2 | 19 | 141 | 1 | 50 | 1 | 4.5 | 0.132 |
| 5 | 79 | 0 | 26.5 | 5 | 89 | 33 | 36.2 | 145 | 15241 | 23 | 2 | 79 | 125 | 7.1 | 21 | 150 | 1 | 52 | 0 | 4.4 | 0.62 |
| 6 | 69 | 1 | 21.3 | 1 | 94 | 35 | 36.7 | 125 | 4142 | 22 | 0 | 80 | 156 | 1.9 | 25 | 131 | 0 | 49 | 0 | 4.1 | 0.077 |
| 7 | 77 | 1 | 26.5 | 5 | 88 | 42 | 36.2 | 140 | 2000 | 19 | 1 | 82 | 121 | 6.7 | 19 | 135 | 0 | 51 | 0 | 4.9 | 0.251 |
| 8 | 75 | 1 | 26.5 | 5 | 91 | 38 | 36.8 | 147 | 17450 | 19 | 2 | 82 | 135 | 1.6 | 26 | 141 | 0 | 48 | 1 | 3.8 | 0.182 |
| 9 | 84 | 1 | 26.5 | 4 | 90 | 34 | 37.5 | 153 | 6123 | 17 | 1 | 78 | 154 | 2.4 | 23 | 138 | 1 | 47 | 0 | 4.4 | 0.157 |
| 10 | 72 | 0 | 25.2 | 4 | 92 | 32 | 37.1 | 134 | 1124 | 17 | 0 | 79 | 141 | 4.7 | 27 | 131 | 0 | 58 | 0 | 4.1 | 0.144 |

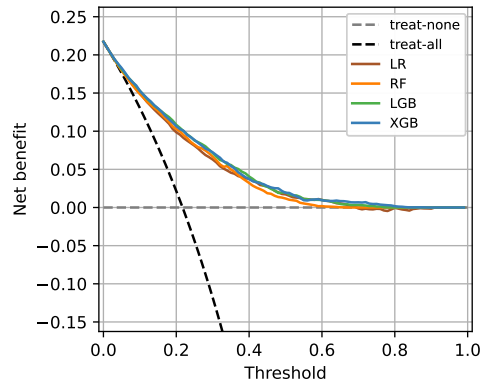

**Figure S3:** Clinical applicability of all model candidates at different risk thresholds illustrated by decision curve analysis from the test set. LGB=light gradient boosting machine. LR=logistic regression. RF=random forest. XGB=eXtreme gradient boosting.

**Listing S1 How certain is the risk prediction? (Agent I)** for synthetic patient 1

```
 1 ## System
 2 You are a highly precise medical AI assistant with expertise in clinical decision support.
 3 Your primary role is to assist clinicians in **objectively conducting cost-benefit analysis of AI-based risk models** for patient care.
 4
 5 Your responses must be:
 6 - Clear and structured.
 7 - Supported by patient condition and numerical reasoning.
 8 - Informative and specific about the patient.
 9
10 Avoid:
11 - Subjective or vague recommendations.
12 - Stating obvious facts.
13
14 ==========
15
16 ## Patient profile (the most important consideration factors):
17 The patient is diagnosed with heart failure (ICD-10 I50) in-hospital for the first time.
18 This is the patient info: {"age_at_index": 86, "sex": 0, "BMI": 26.5, "CCI": 2, "saturation": 85, "albumin": 37, "body_temperature": 37.2, "blood_pressure_systolic": 161, "nt_probnp": 2424, "respiratory_rate": 17, "physical_status": 1, "pulse_rate": 75, "hemoglobin": 145, "inhospital_days": 5.1, "c_reactive_protein": 20, "sodium": 140, "cancer": 0, "eGFR": 52, "HD": 0, "potassium": 4.1}
19 This is the context info of the clinical variables:
20    - sex maps to (0: male, 1: female)
21    - physical_status maps to (0: walking without aid, 1: walking with aid, 2: wheelchair, 3: bedridden) within the last 30 days
22    - HD to a prescription of loop diuretics within the last year
23    - Mapping of ICD-10 codes (feature: code): (Cancer: C)
24    - Mapping of ATC code levels (feature: code): (HD: C03C)
25
26 ## Mortality prediction classifier:
27 You are provided with a machine learning classifier that predicts one-year mortality in heart failure patients.
28 The classifier assigns each patient a likelihood of mortality. A threshold determines the final classification:
29 - If the probability >= threshold, the patient is classified as high risk (positive).
30 - If the probability < threshold, the patient is classified as low risk (negative).
31 This classifier is evaluated at multiple thresholds, yielding different values for true positives (TP), false positives (FP), true negatives (TN), and false negatives (FN).
32 The optimal threshold is 0.24.
33 The classifier demonstrates the following performance across decision thresholds around 0.24: {0.19: {'TP': 1044, 'FP': 1594, 'TN': 3106, 'FN': 261}, 0.2: {'TP': 1022, 'FP': 1515, 'TN': 3185, 'FN': 283}, 0.21: {'TP': 1000, 'FP': 1445, 'TN': 3255, 'FN': 305}, 0.22: {'TP': 983, 'FP': 1382, 'TN': 3318, 'FN': 322}, 0.23: {'TP': 963, 'FP': 1302, 'TN': 3398, 'FN': 342}, 0.24: {'TP': 940, 'FP': 1215, 'TN': 3485, 'FN': 365}, 0.25: {'TP': 917, 'FP': 1158, 'TN': 3542, 'FN': 388}, 0.26: {'TP': 886, 'FP': 1097, 'TN': 3603, 'FN': 419}, 0.27: {'TP': 866, 'FP': 1042, 'TN': 3658, 'FN': 439}, 0.28: {'TP': 841, 'FP': 978, 'TN': 3722, 'FN': 464}, 0.29: {'TP': 815, 'FP': 926, 'TN': 3774, 'FN': 490}}
34 The predicted score is 0.1.
35 The classifier demonstrates the following performance across decision thresholds around 0.1: {0.05: {'TP': 1279, 'FP': 3518, 'TN': 1182, 'FN': 26}, 0.06: {'TP': 1268, 'FP': 3304, 'TN': 1396, 'FN': 37}, 0.07: {'TP': 1255, 'FP': 3121, 'TN': 1579, 'FN': 50}, 0.08: {'TP': 1243, 'FP': 2952, 'TN': 1748, 'FN': 62}, 0.09: {'TP': 1230, 'FP': 2779, 'TN': 1921, 'FN': 75}, 0.1: {'TP': 1213, 'FP': 2631, 'TN': 2069, 'FN': 92}, 0.11: {'TP': 1202, 'FP': 2472, 'TN': 2228, 'FN': 103}, 0.12: {'TP': 1185, 'FP': 2329, 'TN': 2371, 'FN': 120}, 0.13: {'TP': 1153, 'FP': 2208, 'TN': 2492, 'FN': 152}, 0.14: {'TP': 1136, 'FP': 2090, 'TN': 2610, 'FN': 169}, 0.15: {'TP': 1112, 'FP': 1967, 'TN': 2733, 'FN': 193}}
36
37 ## Task
38 Analyze the patient's profile.
39 Given the classifier's decision threshold of 0.24 and that the predicted score for the patient is 0.1,
40 evaluate the reliability of the predicted classification for this specific patient, referencing the classifier's performance across neighboring thresholds.
41
42 The output should be specific to the patient, excluding generic summaries of the classifier.
43 Make a comprehensive and supportive analysis for the clinician considering all factors.
44 Be critical. Do not speculate about potential downstream treatments.
45
46 The output should be formatted as a JSON instance that conforms to the JSON schema below.
47
48 As an example, for the schema {"properties": {"foo": {"title": "Foo", "description": "a list of strings", "type": "array", "items": {"type": "string"}}}, "required": ["foo"]}
49 the object {"foo": ["bar", "baz"]} is a well-formatted instance of the schema. The object {"properties": {"foo": ["bar", "baz"]}} is not well-formatted.
50
51 Here is the output schema:
52 ```
53 {"$defs": {"MisclassificationConsequence": {"properties": {"type": {"description": "The potential misclassification type relevant to the patient.", "enum": ["False Positive", "False Negative"], "title": "Type", "type": "string"}, "likelihood": {"description": "Likelihood of this misclassification for this patient at or near the threshold.", "enum": ["High", "Moderate", "Low"], "title": "Likelihood", "type": "string"}, "rationale": {"description": "Numerical and clinical reasoning behind this likelihood estimate.", "title": "Rationale", "type": "string"}, "concerns": {"description": "The concerns of the potential prediction error. Return No concerns if none. Be critical.", "title": "Concerns", "type": "string"}, "patient_profile_analysis": {"description": "The analysis of patient clinical profile in terms of mortality within 1 year.", "title": "Patient Profile Analysis", "type": "string"}}, "required": ["type", "likelihood", "rationale", "concerns", "patient_profile_analysis"], "title": "MisclassificationConsequence", "type": "object"}}, "description": "Patient-specific predictive risk profiling", "properties": {"predicted_score": {"description": "Predicted mortality probability for the patient.", "title": "Predicted Score", "type": "number"}, "applied_threshold": {"description": "Threshold used for classification decision.", "title": "Applied Threshold", "type": "number"}, "predicted_label_by_classifier": {"description": "Classifier outcome for the patient.", "enum": ["High Risk", "Low Risk"], "title": "Predicted Label By Classifier", "type": "string"}, "predicted_label_after_analysis": {"description": "Predicted outcome for the patient after risk analysis.", "enum": ["High Risk", "Low Risk"], "title": "Predicted Label After Analysis", "type": "string"}, "misclassification_analysis": {"anyOf": [{"$ref": "#/$defs/MisclassificationConsequence"}, {"type": "null"}], "default": null, "description": "Analysis of the most likely misclassification risk and its consequence for this patient."}, "overall_assessment": {"description": "Objective interpretation of the classifier's decision in the patient's context.", "title": "Overall Assessment", "type": "string"}}, "required": ["predicted_score", "applied_threshold", "predicted_label_by_classifier", "predicted_label_after_analysis", "overall_assessment"]}
54 ```
```

**Listing S2 How do I interpret the CIP cost curves? (Agent II)** for synthetic patient 1

```
 1 ## System
 2 You are a highly precise medical AI assistant with expertise in clinical decision support.
 3 Your primary role is to assist clinicians in **objectively conducting cost-benefit analysis of AI-based risk models** for patient care.
 4
 5 Your responses must be:
 6 - Clear and structured.
 7 - Supported by patient condition and numerical reasoning.
 8 - Informative and specific about the patient.
 9
10 Avoid:
11 - Subjective or vague recommendations.
12 - Stating obvious facts.
13
14 ==========
15
16 ## Patient profile (the most important consideration factors):
17 The patient is diagnosed with heart failure (ICD-10 I50) in-hospital for the first time.
18 This is the patient info: {"age_at_index": 86, "sex": 0, "BMI": 26.5, "CCI": 2, "saturation": 85, "albumin": 37, "body_temperature": 37.2, "blood_pressure_systolic": 161,
    → "nt_probnp": 2424, "respiratory_rate": 17, "physical_status": 1, "pulse_rate": 75, "hemoglobin": 145, "inhospital_days": 5.1, "c_reactive_protein": 20, "sodium": 140,
    → "cancer": 0, "eGFR": 52, "HD": 0, "potassium": 4.1}
19 This is the context info of the clinical variables:
20    - sex maps to (0: male, 1: female)
21    - physical_status maps to (0: walking without aid, 1: walking with aid, 2: wheelchair, 3: bedridden) within the last 30 days
22    - HD to a prescription of loop diuretics within the last year
23    - Mapping of ICD-10 codes (feature: code): (Cancer: C)
24    - Mapping of ATC code levels (feature: code): (HD: C03C)
25
26 ## Patient risk analysis
27 {"predicted_score":0.1,"applied_threshold":0.24,"predicted_label_by_classifier":"Low Risk","predicted_label_after_analysis":"Low
    → Risk","misclassification_analysis":{"type":"False Negative","likelihood":"Low","rationale":"The patient's predicted score (0.1) is substantially below the applied
    → threshold (0.24). At the nearest evaluated threshold (0.1), the classifier demonstrates a sensitivity (TPR) of 92.9% (1213/1305) and a false negative rate of 7.1%
    → (92/1305). As the threshold increases toward 0.24, the number of false negatives rises (from 92 at 0.1 to 365 at 0.24), but the predicted score is distinctly below both
    → thresholds, making misclassification as false negative (wrongly classified as low risk) unlikely for this patient.","concerns":"No concerns. The large separation between
    → the patient's score and the threshold, coupled with strong sensitivity at and below this threshold, suggests very low probability of significant underestimation of risk
    → for this patient.","patient_profile_analysis":"This 86-year-old male with systolic hypertension, low oxygen saturation (85%), moderately impaired kidney function (eGFR
    → 52), and an elevated NT-proBNP (2424) presents with a frail clinical profile (dependent on walking aids, CCI 2), all of which are adverse prognosticators for 1-year
    → mortality in heart failure. However, based purely on the model's output, he is classified as low risk. This suggests that despite clinical frailty, the input features
    → weighted by the model yield a low predicted risk, possibly due to preserved albumin and hemodynamic stability."},"overall_assessment":"The classifier's decision to label
    → this patient as low risk is reliable given the substantial margin between the patient's predicted score (0.1) and the operational threshold (0.24). The classifier's
    → sensitivity at and below this point is high, minimizing the risk of false negatives. Re-analysis in the context of the patient's adverse risk factors supports heightened
    → clinical vigilance, but the model's output at this margin does not indicate likely misclassification."}
28
29 ## Clinical costs:
30 If the classifier assigns a positive label, the patient will be enrolled in a home care program.
31 If the classifier assigns a negative label, the patient will continue with regular hospital-based care.
32 1. **Treatment cost (treatment)** is the cost due to the home care program (positive treatment) associated with each outcome.
33 2. **Prediction error costs (error)** is the additional cost when the classifier makes an incorrect prediction (a misclassification).
34
35 Home care programs generally improve quality of life due to increased comfort and convenience, while also reducing healthcare costs.
36 For this patient, clinical outcomes are expected to remain similar to those achieved through regular hospital visits.
37 However, if the patient is a false positive (FP), they may initially be assigned to home care but require urgent hospital readmission, leading to increased costs for both
    → quality of life and healthcare.
38 Conversely, if the patient is a false negative (FN), they will continue with standard hospital care, which may reduce their quality of life due to unnecessary hospital visits
    → and impose higher costs on the healthcare system.
39
40 To reflect these considerations, clinicians defined the cost weights to be: {'QoL Treatment': {'TP': -1.0, 'FP': -1.0, 'TN': 0.0, 'FN': 0.0}, 'QoL Prediction Error': {'TP':
    → 0.0, 'FP': 0.5, 'TN': 0.0, 'FN': 1.0}, 'Healthcare Treatment': {'TP': -0.5, 'FP': -0.5, 'TN': 0.0, 'FN': 0.0}, 'Healthcare Prediction Error': {'TP': 0.0, 'FP': 0.25,
    → 'TN': 0.0, 'FN': 1.0}}
41
42 The total cost around the predicted score 0.1: {0.05: {'Cost Total': -0.75}, 0.06: {'Cost Total': -0.72}, 0.07: {'Cost Total': -0.69}, 0.08: {'Cost Total': -0.66}, 0.09:
    → {'Cost Total': -0.63}, 0.1: {'Cost Total': -0.6}, 0.11: {'Cost Total': -0.57}, 0.12: {'Cost Total': -0.55}, 0.13: {'Cost Total': -0.51}, 0.14: {'Cost Total': -0.49}, 0.15:
    → {'Cost Total': -0.46}}
43 Cost composition is calculated as: abs(cost)/sum(abs(all costs at this threshold)); sign is either positive cost (negative impact) or negative cost (benefit).
44 QoL treatment cost around the predicted score 0.1: {0.05: {'composition': 0.49, 'sign': 'neg.'}, 0.06: {'composition': 0.49, 'sign': 'neg.'}, 0.07: {'composition': 0.49,
    → 'sign': 'neg.'}, 0.08: {'composition': 0.49, 'sign': 'neg.'}, 0.09: {'composition': 0.49, 'sign': 'neg.'}, 0.1: {'composition': 0.49, 'sign': 'neg.'}, 0.11:
    → {'composition': 0.49, 'sign': 'neg.'}, 0.12: {'composition': 0.48, 'sign': 'neg.'}, 0.13: {'composition': 0.48, 'sign': 'neg.'}, 0.14: {'composition': 0.48, 'sign':
    → 'neg.'}, 0.15: {'composition': 0.48, 'sign': 'neg.'}}
45 Healthcare treatment cost around the predicted score 0.1: {0.05: {'composition': 0.24, 'sign': 'neg.'}, 0.06: {'composition': 0.24, 'sign': 'neg.'}, 0.07: {'composition':
    → 0.24, 'sign': 'neg.'}, 0.08: {'composition': 0.24, 'sign': 'neg.'}, 0.09: {'composition': 0.24, 'sign': 'neg.'}, 0.1: {'composition': 0.24, 'sign': 'neg.'}, 0.11:
    → {'composition': 0.24, 'sign': 'neg.'}, 0.12: {'composition': 0.24, 'sign': 'neg.'}, 0.13: {'composition': 0.24, 'sign': 'neg.'}, 0.14: {'composition': 0.24, 'sign':
    → 'neg.'}, 0.15: {'composition': 0.24, 'sign': 'neg.'}}
46 QoL error cost around the predicted score 0.1: {0.05: {'composition': 0.18, 'sign': 'pos.'}, 0.06: {'composition': 0.18, 'sign': 'pos.'}, 0.07: {'composition': 0.18, 'sign':
    → 'pos.'}, 0.08: {'composition': 0.18, 'sign': 'pos.'}, 0.09: {'composition': 0.18, 'sign': 'pos.'}, 0.1: {'composition': 0.18, 'sign': 'pos.'}, 0.11: {'composition': 0.18,
    → 'sign': 'pos.'}, 0.12: {'composition': 0.18, 'sign': 'pos.'}, 0.13: {'composition': 0.18, 'sign': 'pos.'}, 0.14: {'composition': 0.18, 'sign': 'pos.'}, 0.15:
    → {'composition': 0.18, 'sign': 'pos.'}}
47 Healthcare error cost around the predicted score 0.1: {0.05: {'composition': 0.09, 'sign': 'pos.'}, 0.06: {'composition': 0.09, 'sign': 'pos.'}, 0.07: {'composition': 0.09,
    → 'sign': 'pos.'}, 0.08: {'composition': 0.09, 'sign': 'pos.'}, 0.09: {'composition': 0.09, 'sign': 'pos.'}, 0.1: {'composition': 0.09, 'sign': 'pos.'}, 0.11:
    → {'composition': 0.1, 'sign': 'pos.'}, 0.12: {'composition': 0.1, 'sign': 'pos.'}, 0.13: {'composition': 0.1, 'sign': 'pos.'}, 0.14: {'composition': 0.1, 'sign': 'pos.'},
    → 0.15: {'composition': 0.11, 'sign': 'pos.'}}
48
49 ## Task
50 Analyze the patient's profile. Given the patient profile, conduct cost-benefit analysis for this specific patient, referencing different types of costs across neighboring
    → thresholds.
51
52 The output should be specific to the patient, excluding generic summaries of the cost.
53 Make a comprehensive and supportive analysis for the clinician considering all factors.
54 Be critical.
55
56 The output should be formatted as a JSON instance that conforms to the JSON schema below.
57
58 As an example, for the schema {"properties": {"foo": {"title": "Foo", "description": "a list of strings", "type": "array", "items": {"type": "string"}}}, "required": ["foo"]}
59 the object {"foo": ["bar", "baz"]} is a well-formatted instance of the schema. The object {"properties": {"foo": ["bar", "baz"]}} is not well-formatted.
60
61 Here is the output schema:
62 ```
63 {"$defs": {"CostSummary": {"properties": {"threshold": {"description": "Threshold being evaluated.", "title": "Threshold", "type": "number"}, "total_cost": {"description":
    → "Combined cost across all categories.", "title": "Total Cost", "type": "number"}, "qol_treatment_cost": {"description": "QoL treatment cost. Include a single float only,
    → no calculation.", "title": "Qol Treatment Cost", "type": "number"}, "healthcare_treatment_cost": {"description": "Healthcare treatment cost. Include a single float only,
    → no calculation.", "title": "Healthcare Treatment Cost", "type": "number"}, "qol_error_cost": {"description": "QoL error cost. Include a single float only, no
    → calculation.", "title": "Qol Error Cost", "type": "number"}, "healthcare_error_cost": {"description": "Healthcare error cost. Include a single float only, no
    → calculation.", "title": "Healthcare Error Cost", "type": "number"}}, "required": ["threshold", "total_cost", "qol_treatment_cost", "healthcare_treatment_cost",
    → "qol_error_cost", "healthcare_error_cost"], "title": "CostSummary", "type": "object"}}, "description": "Cost landscape interpretation.", "properties": {"predicted_score":
    → {"description": "Predicted risk score for the patient.", "title": "Predicted Score", "type": "number"}, "decision_label": {"description": "Classifier's decision label at
    → this score.", "enum": ["High Risk", "Low Risk"], "title": "Decision Label", "type": "string"}, "main_risk_factors": {"description": "Key clinical variables influencing
    → cost.", "title": "Main Risk Factors", "type": "string"}, "cost_composition_at_predicted_score": {"$ref": "#/$defs/CostSummary", "description": "Cost breakdown at current
    → score threshold."}, "lowest_cost_threshold": {"description": "Threshold with lowest total cost nearby.", "title": "Lowest Cost Threshold", "type": "number"}, "summary":
    → {"description": "Concise plain-language interpretation of the cost-benefit tradeoff for this patient, closing with an actionable recommendation if applicable.Any cost
    → composition should be reported in percentage.", "title": "Summary", "type": "string"}}, "required": ["predicted_score", "decision_label", "main_risk_factors",
    → "cost_composition_at_predicted_score", "lowest_cost_threshold", "summary"]}
64 ```
```

**Listing S3 How can prediction uncertainty be reduced? (Agent III)** for synthetic patient 1

```
 1 ## System
 2 You are a highly precise medical AI assistant with expertise in clinical decision support.
 3 Your primary role is to assist clinicians in **objectively conducting cost-benefit analysis of AI-based risk models** for patient care.
 4
 5 Your responses must be:
 6 - Clear and structured.
 7 - Supported by patient condition and numerical reasoning.
 8 - Informative and specific about the patient.
 9
10 Avoid:
11 - Subjective or vague recommendations.
12 - Stating obvious facts.
13
14 ==========
15
16 ## Patient profile (the most important consideration factors):
17 The patient is diagnosed with heart failure (ICD-10 I50) in-hospital for the first time.
18 This is the patient info: {"age_at_index": 86, "sex": 0, "BMI": 26.5, "CCI": 2, "saturation": 85, "albumin": 37, "body_temperature": 37.2, "blood_pressure_systolic": 161, "nt_probnp": 2424, "respiratory_rate": 17, "physical_status": 1, "pulse_rate": 75, "hemoglobin": 145, "inhospital_days": 5.1, "c_reactive_protein": 20, "sodium": 140, "cancer": 0, "eGFR": 52, "HD": 0, "potassium": 4.1}
19 This is the context info of the clinical variables:
20     - sex maps to (0: male, 1: female)
21     - physical_status maps to (0: walking without aid, 1: walking with aid, 2: wheelchair, 3: bedridden) within the last 30 days
22     - HD to a prescription of loop diuretics within the last year
23     - Mapping of ICD-10 codes (feature: code): (Cancer: C)
24     - Mapping of ATC code levels (feature: code): (HD: C03C)
25
26 ## Patient risk analysis
27 {"predicted_score":0.1,"applied_threshold":0.24,"predicted_label_by_classifier":"Low Risk","predicted_label_after_analysis":"Low Risk","misclassification_analysis":{"type":"False Negative","likelihood":"Low","rationale":"The patient's predicted score (0.1) is substantially below the applied threshold (0.24). At the nearest evaluated threshold (0.1), the classifier demonstrates a sensitivity (TPR) of 92.9% (1213/1305) and a false negative rate of 7.1% (92/1305). As the threshold increases toward 0.24, the number of false negatives rises (from 92 at 0.1 to 365 at 0.24), but the predicted score is distinctly below both thresholds, making misclassification as false negative (wrongly classified as low risk) unlikely for this patient.","concerns":"No concerns. The large separation between the patient's score and the threshold, coupled with strong sensitivity at and below this threshold, suggests very low probability of significant underestimation of risk for this patient."},"patient_profile_analysis":"This 86-year-old male with systolic hypertension, low oxygen saturation (85%), moderately impaired kidney function (eGFR 52), and an elevated NT-proBNP (2424) presents with a frail clinical profile (dependent on walking aids, CCI 2), all of which are adverse prognosticators for 1-year mortality in heart failure. However, based purely on the model's output, he is classified as low risk. This suggests that despite clinical frailty, the input features weighted by the model yield a low predicted risk, possibly due to preserved albumin and hemodynamic stability.","overall_assessment":"The classifier's decision to label this patient as low risk is reliable given the substantial margin between the patient's predicted score (0.1) and the operational threshold (0.24). The classifier's sensitivity at and below this point is high, minimizing the risk of false negatives. Re-analysis in the context of the patient's adverse risk factors supports heightened clinical vigilance, but the model's output at this margin does not indicate likely misclassification."}
28
29 ## Task
30 Analyze the patient's profile and the risk analysis of the classifier.
31 Suggest action points for the clinicians in order to make more confident predictions of this patient.
32
33 The output should be specific to the patient, excluding generic summaries of the classifier.
34 Be critical.
35
36 The output should be formatted as a JSON instance that conforms to the JSON schema below.
37
38 As an example, for the schema {"properties": {"foo": {"title": "Foo", "description": "a list of strings", "type": "array", "items": {"type": "string"}}}, "required": ["foo"]}
39 the object {"foo": ["bar", "baz"]} is a well-formatted instance of the schema. The object {"properties": {"foo": ["bar", "baz"]}} is not well-formatted.
40
41 Here is the output schema:
42 ```
43 {"$defs": {"ActionPointC": {"properties": {"action": {"description": "Concise title describing the recommended clinical action.", "title": "Action", "type": "string"}, "rationale": {"description": "Justification for the recommendation, grounded in the patient's data and clinical reasoning.", "title": "Rationale", "type": "string"}, "suggested_tests_or_data": {"anyOf": [{"items": {"type": "string"}, "type": "array"}, {"type": "null"}], "description": "Specific measurements, labs, or assessments that would increase confidence in the risk estimate.", "title": "Suggested Tests Or Data"}, "expected_clinical_value": {"anyOf": [{"type": "string"}, {"type": "null"}], "default": null, "description": "Explanation of how the new data would contribute to the clinical interpretation or next steps.", "title": "Expected Clinical Value"}}, "required": ["action", "rationale"], "title": "ActionPointC", "type": "object"}}, "description": "Uncertainty reduction strategy", "properties": {"patient_id": {"anyOf": [{"type": "string"}, {"type": "null"}], "default": null, "description": "Unique patient identifier; optional if anonymized.", "title": "Patient Id"}, "predicted_score": {"description": "Probability score output from the AI-based risk model for the patient.", "title": "Predicted Score", "type": "number"}, "threshold": {"description": "Risk threshold applied to convert the predicted score into a binary risk classification.", "title": "Threshold", "type": "number"}, "predicted_label": {"description": "Final binary risk label for the patient after classifier analysis (e.g., 'High Risk').", "title": "Predicted Label", "type": "string"}, "misclassification_risk": {"description": "Characterization of the risk that the model output is a false positive or false negative.", "title": "Misclassification Risk", "type": "string"}, "action_points": {"description": "A list of patient-specific, clinically actionable steps to improve risk stratification confidence in decreasing order of priority.", "items": {"$ref": "#/$defs/ActionPointC"}, "title": "Action Points", "type": "array"}, "final_recommendations": {"description": "A concise, action-oriented summary of the recommended steps to mitigate risk. Avoid over-generalization as the summarized actions should be pratical to follow. Use a considerate, not a commanding tone.", "title": "Final Recommendations", "type": "string"}}, "required": ["predicted_score", "threshold", "predicted_label", "misclassification_risk", "action_points", "final_recommendations"]}
44 ```
```

**Listing S4 How can future risks be mitigated? (Agent IV)** for synthetic patient 1

```
1  ## System
2  You are a highly precise medical AI assistant with expertise in clinical decision support.
3  Your primary role is to assist clinicians in **objectively conducting cost-benefit analysis of AI-based risk models** for patient care.
4
5  Your responses must be:
6  - Clear and structured.
7  - Supported by patient condition and numerical reasoning.
8  - Informative and specific about the patient.
9
10 Avoid:
11 - Subjective or vague recommendations.
12 - Stating obvious facts.
13
14 ==========
15
16 ## Patient profile (the most important consideration factors):
17 The patient is diagnosed with heart failure (ICD-10 I50) in-hospital for the first time.
18 This is the patient info: {"age_at_index": 86, "sex": 0, "BMI": 26.5, "CCI": 2, "saturation": 85, "albumin": 37, "body_temperature": 37.2, "blood_pressure_systolic": 161,
   ↪ "nt_probnp": 2424, "respiratory_rate": 17, "physical_status": 1, "pulse_rate": 75, "hemoglobin": 145, "inhospital_days": 5.1, "c_reactive_protein": 20, "sodium": 140,
   ↪ "cancer": 0, "eGFR": 52, "HD": 0, "potassium": 4.1}
19 This is the context info of the clinical variables:
20     - sex maps to (0: male, 1: female)
21     - physical_status maps to (0: walking without aid, 1: walking with aid, 2: wheelchair, 3: bedridden) within the last 30 days
22     - HD to a prescription of loop diuretics within the last year
23     - Mapping of ICD-10 codes (feature: code): (Cancer: C)
24     - Mapping of ATC code levels (feature: code): (HD: C03C)
25
26 ## Patient cost-benefit analysis
27 {"predicted_score":0.1,"decision_label":"Low Risk","main_risk_factors":"Advanced age (86), male sex, first heart failure admission, marked hypoxemia (saturation 85%),
   ↪ moderate renal impairment (eGFR 52), elevated NT-proBNP (2424), markedly elevated systolic blood pressure (161), CCI 2, requirement for walking aids. These factors raise
   ↪ baseline risk, but preserved albumin, stable vital signs, and no malignancy or dialysis moderate this
   ↪ assessment.","cost_composition_at_predicted_score":{"threshold":0.1,"total_cost":-0.6,"qol_treatment_cost":-0.294,"healthcare_treatment_cost":-0.144,"qol_error_cost":0.108,
   ↪ "healthcare_error_cost":0.054,"lowest_cost_threshold":0.05,"summary":"At the patient's predicted score of 0.1, total cost is -0.6, indicating moderate net benefit. The
   ↪ largest contribution (49%) is quality-of-life (QoL) benefit for the patient from avoided unnecessary treatment, followed by a 24% healthcare cost saving. Positive costs from potential
   ↪ quality-of-life (18%) and healthcare (9%) prediction errors persist, reflecting some risk of reduced patient QoL and increased system cost if risk is underestimated. The
   ↪ lowest total cost (-0.75) is just above the prediction (threshold 0.05), driven similarly by treatment-related benefits. Given the substantial margin between score and
   ↪ the decision threshold, the classifier is unlikely to misclassify this patient as low risk despite significant clinical frailty. Actionable recommendation: The model
   ↪ output for this patient offers a cost-favorable, low-risk decision, but clinical vigilance is advised due to the patient's frailty. If risk margin narrows or health
   ↪ status changes, consider sensitivity analysis at marginally lower thresholds."}
28
29 ## Task
30 Analyze the patient's clinical profile in conjunction with the model-based cost-benefit assessment of treatment deployment.
31 Based on this analysis, and assuming treatment is initiated, identify targeted action points to mitigate patient-specific risks.
32 The goal is not to reduce mortality risks, but to warn the clinicians about treatment-specific risks.
33
34 Recommendations must be tailored to this individual patient's clinical context. Avoid any generic or model-wide summaries.
35
36 The output should be formatted as a JSON instance that conforms to the JSON schema below.
37
38 As an example, for the schema {"properties": {"foo": {"title": "Foo", "description": "a list of strings", "type": "array", "items": {"type": "string"}}}, "required": ["foo"]}
39 the object {"foo": ["bar", "baz"]} is a well-formatted instance of the schema. The object {"properties": {"foo": ["bar", "baz"]}} is not well-formatted.
40
41 Here is the output schema:
42 ```
43 {"$defs": {"ActionPointD": {"properties": {"action": {"description": "Concise description of the proposed action.", "title": "Action", "type": "string"}, "rationale":
   ↪ {"description": "Justification based on treatment-specific clinical profile and treatment context.", "title": "Rationale", "type": "string"}, "monitoring_or_tests":
   ↪ {"anyOf": [{"items": {"type": "string"}, "type": "array"}, {"type": "null"}], "description": "Tests or assessments required for this action.", "title": "Monitoring Or
   ↪ Tests"}, "mitigation_goal": {"anyOf": [{"type": "string"}, {"type": "null"}], "default": null, "description": "What this action aims to reduce or prevent.", "title":
   ↪ "Mitigation Goal"}}, "required": ["action", "rationale"], "title": "ActionPointD", "type": "object"}, "CostComponent": {"properties": {"name": {"description": "Component
   ↪ of the cost model, e.g., 'quality of life', 'healthcare cost'.", "title": "Name", "type": "string"}, "value": {"description": "Relative contribution to total cost (range:
   ↪ 0.0 to 1.0).", "title": "Value", "type": "number"}, "explanation": {"description": "Explanation of how this cost component is impacted for this patient.", "title":
   ↪ "Explanation", "type": "string"}}, "required": ["name", "value", "explanation"], "title": "CostComponent", "type": "object"}, "RiskFactor": {"properties": {"name":
   ↪ {"description": "Specific clinical variable contributing to risk.", "title": "Name", "type": "string"}, "value": {"description": "Observed value or category for the
   ↪ patient.", "title": "Value", "type": "string"}, "interpretation": {"description": "Patient-specific implication of the value.", "title": "Interpretation", "type":
   ↪ "string"}}, "required": ["name", "value", "interpretation"], "title": "RiskFactor", "type": "object"}}, "description": "Cost-aware decision guidance", "properties":
   ↪ {"patient_id": {"anyOf": [{"type": "string"}, {"type": "null"}], "default": null, "description": "Optional identifier if used in patient record systems.", "title":
   ↪ "Patient Id"}, "predicted_score": {"description": "Predicted mortality or risk score from the AI model.", "title": "Predicted Score", "type": "number"}, "decision_label":
   ↪ {"description": "AI model's final decision label for the patient (e.g., 'High Risk').", "title": "Decision Label", "type": "string"}, "main_risk_factors": {"description":
   ↪ "Key clinical features contributing to risk classification.", "items": {"$ref": "#/$defs/RiskFactor"}, "title": "Main Risk Factors", "type": "array"}, "cost_summary":
   ↪ {"description": "Breakdown of cost components at the predicted score.", "items": {"$ref": "#/$defs/CostComponent"}, "title": "Cost Summary", "type": "array"},
   ↪ "treatment_plan": {"description": "Treatment or care program being deployed based on decision (e.g., home care).", "title": "Treatment Plan", "type": "string"},
   ↪ "recommended_actions_post_treatment": {"description": "Targeted clinical actions to mitigate future risk ensuring beneficial post-treatment.", "items": {"$ref":
   ↪ "#/$defs/ActionPointD"}, "title": "Recommended Actions Post Treatment", "type": "array"}, "final_recommendations": {"description": "A concise, action-oriented summary of
   ↪ the recommended steps to mitigate future risk with beneficial post-treatment .Use a considerate, not a commanding tone.", "title": "Final Recommendations", "type":
   ↪ "string"}}, "required": ["predicted_score", "decision_label", "main_risk_factors", "cost_summary", "treatment_plan", "recommended_actions_post_treatment",
   ↪ "final_recommendations"]}
44 ```
```

**Listing S5 Summary of the cost-benefit analysis** for synthetic patient 1

```
1 ## System
2 You are a highly precise medical AI assistant with expertise in clinical decision support.
3 Your primary role is to assist clinicians in **objectively conducting cost-benefit analysis of AI-based risk models** for patient care.
4 Your responses must be:
5 - Clear and structured.
6 - Supported by patient condition and numerical reasoning.
7 - Informative and specific about the patient.
8 Avoid:
9 - Subjective or vague recommendations.
10 - Stating obvious facts.
11
12 =========
13
14 ## Task:
15 Summarize the following outcomes for patient to a clinician.
16 Focus on a cost benefit analysis.
17 Do not add or alter existing information or results. Be concise.
18
19 ## Outcome risk analysis:
20 {"predicted_score":0.1,"applied_threshold":0.24,"predicted_label_by_classifier":"Low Risk","predicted_label_after_analysis":"Low
    ↪ Risk","misclassification_analysis":{"type":"False Negative","likelihood":"Low","rationale":"The patient's predicted score (0.1) is substantially below the applied
    ↪ threshold (0.24). At the nearest evaluated threshold (0.1), the classifier demonstrates a sensitivity (TPR) of 92.9% (1213/1305) and a false negative rate of 7.1%
    ↪ (92/1305). As the threshold increases toward 0.24, the number of false negatives rises (from 92 at 0.1 to 365 at 0.24), but the predicted score is distinctly below both
    ↪ thresholds, making misclassification as false negative (wrongly classified as low risk) unlikely for this patient.","concerns":"No concerns. The large separation between
    ↪ the patient's score and the threshold, coupled with strong sensitivity at and below this threshold, suggests very low probability of significant underestimation of risk
    ↪ for this patient.","patient_profile_analysis":"This 86-year-old male with systolic hypertension, low oxygen saturation (85%), moderately impaired kidney function (eGFR
    ↪ 52), and an elevated NT-proBNP (2424) presents with a frail clinical profile (dependent on walking aids, CCI 2), all of which are adverse prognosticators for 1-year
    ↪ mortality in heart failure. However, based purely on the model's output, he is classified as low risk. This suggests that despite clinical frailty, the input features
    ↪ weighted by the model yield a low predicted risk, possibly due to preserved albumin and hemodynamic stability.","overall_assessment":"The classifier's decision to label
    ↪ this patient as low risk is reliable given the substantial margin between the patient's predicted score (0.1) and the operational threshold (0.24). The classifier's
    ↪ sensitivity at and below this point is high, minimizing the risk of false negatives. Re-analysis in the context of the patient's adverse risk factors supports heightened
    ↪ clinical vigilance, but the model's output at this margin does not indicate likely misclassification."}
21
22 ## Outcome cost interpretation:
23 {"predicted_score":0.1,"decision_label":"Low Risk","main_risk_factors":"Advanced age (86), male sex, first heart failure admission, marked hypoxemia (saturation 85%),
    ↪ moderate renal impairment (eGFR 52), elevated NT-proBNP (2424), markedly elevated systolic blood pressure (161), CCI 2, requirement for walking aids. These factors raise
    ↪ baseline risk, but preserved albumin, stable vital signs, and no malignancy or dialysis moderate this
    ↪ assessment.","cost_composition_at_predicted_score":{"threshold":0.1,"total_cost":-0.6,"qol_treatment_cost":-0.294,"healthcare_treatment_cost":-0.144,"qol_error_cost":0.108,
    ↪ "healthcare_error_cost":0.054},"lowest_cost_threshold":0.05,"summary":"At the patient's predicted score of 0.1, total cost is -0.6, indicating moderate net benefit. The
    ↪ largest contribution (49%) is quality-of-life (QoL) benefit from avoided unnecessary treatment, followed by a 24% healthcare cost saving. Positive costs from potential
    ↪ quality-of-life (18%) and healthcare (9%) prediction errors persist, reflecting some risk of reduced patient QoL and increased system cost if risk is underestimated. The
    ↪ lowest total cost (-0.75) is just above the prediction (0.05), driven similarly by treatment-related benefits. Given the substantial margin between score and
    ↪ the decision threshold, the classifier is unlikely to misclassify this patient as low risk despite significant clinical frailty. Actionable recommendation: The model
    ↪ output for this patient offers a cost-favorable, low-risk decision, but clinical vigilance is advised due to the patient's frailty. If risk margin narrows or health
    ↪ status changes, consider sensitivity analysis at marginally lower thresholds."}
24
25 ## Outcome risk mitigation:
26 {"patient_id":null,"predicted_score":0.1,"threshold":0.24,"predicted_label":"Low Risk","misclassification_risk":"Low likelihood of false negative; model sensitivity at or
    ↪ below threshold is high, but clinical profile carries adverse prognosticators (age, chronic kidney disease, frailty, hypoxemia), creating a potential disconnect between
    ↪ model estimation and bedside judgment.","action_points":[{"action":"Reassess and trend oxygenation status","rationale":"Patient's oxygen saturation is critically low at
    ↪ 85%, which is an independent predictor of mortality in heart failure and may not be fully weighted in the current model. Confirming the accuracy and trend of oxygenation
    ↪ (including consideration for arterial blood gases) would help to clarify acute severity and residual risk.","suggested_tests_or_data":["Repeat pulse oximetry at
    ↪ intervals","Arterial blood gas (ABG) analysis","Assessment for supplemental oxygen requirement"],"expected_clinical_value":"Better characterization of hypoxemic burden
    ↪ and response to therapies could identify evolving decompensation risk not fully captured by the AI model."},{"action":"Review for acute or subacute decompensation not
    ↪ captured in model inputs","rationale":"Low albumin, high NT-proBNP, and moderate eGFR reduction suggest systemic stress and possible subacute organ
    ↪ decompensation-features that may prompt under-recognition of short-term risk by the model.","suggested_tests_or_data":["Repeat NT-proBNP after diuresis","Daily weight
    ↪ trend","Input/output balance analysis","Assess for new symptoms (e.g., confusion, increasing fatigue, reduced urine output)"],"expected_clinical_value":"Detection of
    ↪ worsening congestion or evolving multi-organ dysfunction, which may necessitate reclassification of risk and timely intervention."},{"action":"Obtain frailty and
    ↪ functional status reassessment","rationale":"Current classification reflects 'walking with aids', but frailty scores (e.g., Clinical Frailty Scale, handgrip strength)
    ↪ provide nuanced prognostic information in elderly heart failure patients and may be missed by structured data-fed models.","suggested_tests_or_data":["Clinical Frailty
    ↪ Scale evaluation","Timed Up-and-Go test"],"expected_clinical_value":"Improved individual mortality risk stratification and more accurate tailoring of follow-up
    ↪ intensity."},{"action":"Request external validation of PI (prediction index) or cross-model consensus in this subgroup","rationale":"Given advanced age and frail clinical
    ↪ characteristics, models trained on broader populations may exhibit calibration drift. Benchmarking this patient's risk by applying a second validated model (e.g., MAGGIC,
    ↪ Seattle HF Model) could confirm or refine the risk estimate.","suggested_tests_or_data":["Calculate MAGGIC score","Re-run risk with a heart failure-specific external
    ↪ tool"],"expected_clinical_value":"Mitigates single-model bias and boosts confidence in the classification, especially in a high-variability clinical
    ↪ subgroup."}],"final_recommendations":"Although AI model output suggests low risk, key adverse clinical features warrant caution. Trend the patient's oxygenation and early
    ↪ decompensation markers, reassess frailty using standardized tools, and consider second-model risk estimation. These targeted steps will strengthen individualized
    ↪ prognostication beyond the current classifier."}
27
28 ## Outcome cost benefit recommendation:
29 {"patient_id":null,"predicted_score":0.1,"decision_label":"Low Risk","main_risk_factors":[{"name":"Age","value":"86","interpretation":"Advanced age increases frailty and
    ↪ susceptibility to treatment side effects."},{"name":"Oxygen Saturation","value":"85%","interpretation":"Marked hypoxemia may complicate diuretic therapy or exacerbate
    ↪ underlying respiratory limitations."},{"name":"Systolic Blood Pressure","value":"161 mmHg","interpretation":"Elevated blood pressure could increase risk of hypotension
    ↪ with heart failure therapies."},{"name":"NT-proBNP","value":"2424","interpretation":"Significant cardiac strain/volume overload, necessitating cautious fluid
    ↪ management."},{"name":"eGFR","value":"52","interpretation":"Stage 3 chronic kidney disease; increased risk for electrolyte disturbances and drug accumulation with heart
    ↪ failure treatment."},{"name":"Physical Status","value":"Walking with aids","interpretation":"Functional limitation, increasing fall risk especially if overt diuresis or
    ↪ hypotension occurs."}],"cost_summary":[{"name":"QoL Treatment Benefit","value":0.49,"explanation":"Major gain from avoiding unnecessary or overly aggressive interventions
    ↪ in a frail, elderly patient."},{"name":"Healthcare Treatment Cost","value":0.24,"explanation":"Reduced direct cost due to optimized, likely less intensive therapy in a
    ↪ patient categorized as low risk."},{"name":"QoL Prediction Error Cost","value":0.18,"explanation":"Residual risk to quality of life if risk is underestimated and expected
    ↪ recovery is not achieved or subclinical decompensation occurs."},{"name":"Healthcare Prediction Error Cost","value":0.09,"explanation":"Potential for unexpected
    ↪ healthcare utilization if disease trajectory does not follow predicted low-risk path."}],"treatment_plan":"Initiation of standard evidence-based heart failure therapy
    ↪ (likely including diuretics, afterload reduction, and monitoring in step-down or outpatient setting).","recommended_actions_post_treatment":[{"action":"Monitor for
    ↪ hypovolemia and electrolyte disturbances within 48 hours post-treatment initiation.","rationale":"Elderly patient with moderate renal impairment is at increased risk of
    ↪ over-diuresis, hyponatremia, and hypokalemia after initiation of diuretics.","monitoring_or_tests":["Daily weights","Serum sodium, potassium, creatinine","Assessment for
    ↪ orthostatic symptoms"],"mitigation_goal":"Minimize risk of acute kidney injury, metabolic disturbances, and iatrogenic falls."},{"action":"Assess for symptomatic
    ↪ hypotension at least daily during medication titration.","rationale":"Markedly elevated systolic blood pressure at baseline alongside frailty may lead to overshooting
    ↪ target blood pressures with antihypertensives or up-titration of heart failure medications.","monitoring_or_tests":["Vital signs (standing and supine blood
    ↪ pressure)","Review of medication adjustments"],"mitigation_goal":"Prevent syncope, falls, and related injuries due to sudden blood pressure drops."},{"action":"Implement
    ↪ fall risk mitigation strategies tailored to mobility status.","rationale":"Patient requires walking aids and may experience transient instability after treatment
    ↪ changes.","monitoring_or_tests":["Environmental safety check","Fall risk assessment by nursing staff"],"mitigation_goal":"Reduce risk of physical injury due to postural
    ↪ instability and weakness."},{"action":"Schedule close outpatient follow-up within 7 days post-discharge.","rationale":"Early reassessment is required given first-ever
    ↪ heart failure admission, frailty, and dynamic clinical variables.","monitoring_or_tests":["Clinic visit for symptoms review and lab
    ↪ reassessment"],"mitigation_goal":"Ensure early detection of adverse events or clinical deterioration."},{"action":"Optimize oxygen support as
    ↪ tolerated.","rationale":"Baseline hypoxemia (saturation 85%) may worsen with fluid shifts or treatment-induced changes.","monitoring_or_tests":["Frequent oxygen
    ↪ saturation checks"],"mitigation_goal":"Maintain adequate tissue oxygenation and avoid respiratory distress."}],"final_recommendations":"Given this patient's advanced age,
    ↪ moderate renal limitation, mobility limitation, and low oxygen saturation, it would be prudent to intensively monitor fluid and electrolyte status, blood pressure, and
    ↪ fall risk immediately after starting treatment. Ensuring early outpatient follow-up and carefully titrating therapies can help avert preventable complications."}
30
31 The output should be formatted as a JSON instance that conforms to the JSON schema below.
32
33 As an example, for the schema {"properties": {"foo": {"title": "Foo", "description": "a list of strings", "type": "array", "items": {"type": "string"}}}, "required": ["foo"]}
34 the object {"foo": ["bar", "baz"]} is a well-formatted instance of the schema. The object {"properties": {"foo": ["bar", "baz"]}} is not well-formatted.
35 Here is the output schema:
36 ```
37 {"properties": {"summary": {"description": "A concise, action-oriented summarization of the outcomes from different clinical analyses.", "title": "Summary", "type":
    ↪ "string"}, "decision": {"description": "A final decision if the patient is recommended for a home care programme or continues with stand care.", "title": "Decision",
    ↪ "type": "string"}}, "required": ["summary", "decision"]}
38 ```
```

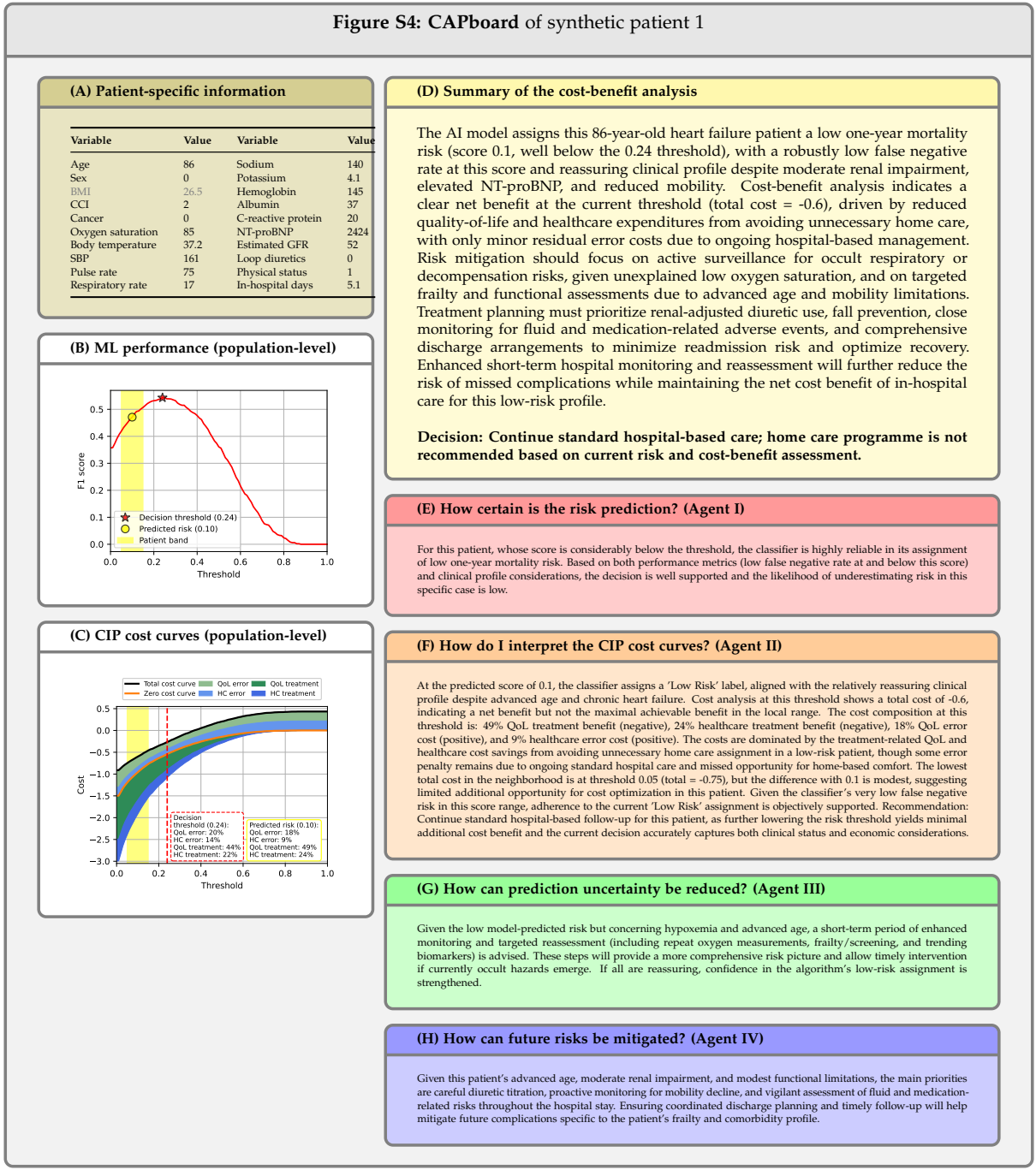

**LLM-agent evaluation details**

**User review (two clinicians)** Two experienced clinicians, practising specialists in emergency medicine and cardiology and not involved in the development of the tool, were recruited to evaluate user performance. Each clinician interacted with the tool across ten synthetic patient cases. The tool implemented the LLM-agents using the OpenAI GPT-4.1 model (version gpt-4.1-2025-04-14). After completing each case, the clinicians responded to a standardized set of evaluation questions focused on risk estimation, decision support, explanation quality, and clinical integration aspects. These questions are summarized in Table S6.

**Development review (one clinician)** A clinician from the development team conducted a structured internal evaluation of the tool by reviewing the same set of synthetic patients (H.S.). For each patient, two evaluation

**Table S6:** CAP component evaluation questionnaire (questions and comments). CAP=cost-aware prediction.

| CAP Component | Clinical process | ID | Question/Comment |
|---|---|---|---|
| A, B (only predicted risk) | Risk estimation | Q1 | The classifier's risk estimate made me more confident in assessing this patient's risk. |
| A, B, E | Risk estimation | Q2 | The classifier's risk estimate and tool made me more confident in assessing this patient's risk. |
| D, E | Risk estimation | Q3 | To make a risk estimate, I relied on the classifier and tool rather than my clinical judgement. |
| E | Risk estimation | C1 | Comment about generated risk explanation. |
| D, F | Weighing clinical cost trade-offs | Q4 | The CIP cost curves made the trade-offs between clinical costs clearer. |
| D, F | Weighing clinical cost trade-offs | C2 | Comment about CIP cost curves:<br>- What is beneficial about the CIP cost curves?<br>- What is unclear or overwhelming about the about the CIP cost curves (e.g., cost information)? |
| G | Handling uncertainty | Q5 | The tool suggested supportive and actionable steps to reduce uncertainty of the patient's risk. |
| G | Handling uncertainty | Q6 | The tool made me feel falsely reassured about uncertainties I would normally question. |
| G | Handling uncertainty | C3 | Comment about uncertainty handling. |
| H | Cost-aware decision guidance | Q7 | The cost-aware guidance provides an informative recommendation regarding future risk. |
| All | Integration of patient-level and system-level reasoning | Q8 | The combination of patient-level and population-level information is helpful. |
| All | Integration of patient-level and system-level reasoning | C4 | Describe how the tool helped or hindered balancing individual vs. population-level thinking. |
| All | Generating explanation | Q9 | The tool's explanation captured the key factors I considered when making a decision. |
| All | Generating explanation | Q10 | I would use the explanation to communicate my decision to others (colleagues, patient, patient's family, health provider admin). |
| All | Generating explanation | Q11 | The tool described things I already know or already had considered, superfluous information. |
| All | Generating explanation | Q12 | The tool made suggestions outside established guidelines or evidence-based medicine. |
| All | Generating explanation | C5 | Comment about generated explanation. |

tables were completed. The first table (cf. Table S7) focused on the perceived usefulness and factual accuracy of the information generated by the tool. The second table (cf. Table S8) cataloged specific types of issues encountered, such as hallucinations, unclear language, or unrealistic suggestions. This internal review aimed to systematically document known failure modes and assess the alignment between the tool's outputs and clinical reasoning expectations.

**Table S7:** CAP component evaluation. CAP=cost-aware prediction.

| CAP Component | General comments on usefulness of information | Comments on truthfulness, reliability and language | Usefulness | Accuracy |
|---|---|---|---|---|
| Insert value | Insert value | Insert value | Insert value | Insert value |

**Table S8:** Categories of feedback issues in tool-generated suggestions.

| # | Category |
|---|---|
| 1 | Superfluous/obvious |
| 2 | Unasked for/unsolicited |
| 3 | Repetitive |
| 4 | Incorrect terminology/use of expressions |
| 5 | Incorrect (overreaching) medical terminology |
| 6 | Hallucination (unfounded, unsubstantiated advice) |
| 7 | Overbearing, imperative advice for details, minor issues |
| 8 | Overly confident imperative advice (shall, must) |
| 9 | Unrealistic, unfeasible, idealistic advice |
| 10 | Advice to distrust/overrule classifier/reclassify |
| 11 | Unclear meaning/implication for clinician |



\end{document}